\documentclass{article}

\PassOptionsToPackage{numbers, sort}{natbib}
\usepackage[preprint]{neurips_2026}

\usepackage[utf8]{inputenc} 
\usepackage[T1]{fontenc}    
\usepackage{hyperref}       
\usepackage{url}            

\usepackage{amsfonts}       
\usepackage{amsmath}        
\usepackage{amssymb}        
\usepackage{amsthm}         
\usepackage{mathtools}      
\usepackage{algorithm}      
\usepackage{algpseudocode}  
\usepackage{nicefrac}       
\usepackage{microtype}      
\usepackage{xcolor}         
\usepackage{enumitem}       
\newcommand{\eg}{\emph{e.g.}}
\newcommand{\ie}{\emph{i.e.}}
\usepackage{graphicx}      
\usepackage{multirow}
\usepackage{booktabs}
\usepackage{caption}
\usepackage{adjustbox}
\usepackage{subcaption}
\usepackage[table]{xcolor}
\usepackage{wrapfig}

\usepackage{aliascnt}

\theoremstyle{plain}
\newtheorem{theorem}{Theorem}
\newaliascnt{proposition}{theorem}\newtheorem{proposition}[proposition]{Proposition}\aliascntresetthe{proposition}
\newaliascnt{lemma}{theorem}\newtheorem{lemma}[lemma]{Lemma}\aliascntresetthe{lemma}
\newaliascnt{corollary}{theorem}\newtheorem{corollary}[corollary]{Corollary}\aliascntresetthe{corollary}
\theoremstyle{definition}
\newaliascnt{definition}{theorem}\newtheorem{definition}[definition]{Definition}\aliascntresetthe{definition}
\newaliascnt{assumption}{theorem}\newtheorem{assumption}[assumption]{Assumption}\aliascntresetthe{assumption}
\theoremstyle{remark}
\newaliascnt{remark}{theorem}\newtheorem{remark}[remark]{Remark}\aliascntresetthe{remark}

\usepackage{cleveref}
\usepackage[most]{tcolorbox}
\crefname{assumption}{Assumption}{Assumptions}
\Crefname{assumption}{Assumption}{Assumptions}
\crefname{theorem}{Theorem}{Theorems}
\Crefname{theorem}{Theorem}{Theorems}
\crefname{proposition}{Proposition}{Propositions}
\Crefname{proposition}{Proposition}{Propositions}
\crefname{lemma}{Lemma}{Lemmas}
\Crefname{lemma}{Lemma}{Lemmas}
\crefname{corollary}{Corollary}{Corollaries}
\Crefname{corollary}{Corollary}{Corollaries}
\crefname{definition}{Definition}{Definitions}
\Crefname{definition}{Definition}{Definitions}
\crefname{remark}{Remark}{Remarks}
\Crefname{remark}{Remark}{Remarks}
\crefname{section}{Appendix}{Appendices}
\Crefname{section}{Appendix}{Appendices}

\DeclareMathOperator{\sg}{sg}
\DeclareMathOperator{\FFT}{\mathcal{F}}
\DeclareMathOperator{\diag}{diag}
\DeclareMathOperator{\PSD}{S}
\DeclareMathOperator*{\argmin}{arg\,min}

\newcommand{\R}{\mathbb{R}}
\newcommand{\C}{\mathbb{C}}
\newcommand{\E}{\mathbb{E}}
\newcommand{\N}{\mathcal{N}}
\renewcommand{\d}{\,\mathrm{d}}
\newcommand{\X}{\mathcal{X}}
\newcommand{\Latents}{\mathcal{Z}}
\newcommand{\Lcrit}{\mathcal{L}_{\mathrm{crit}}}
\newcommand{\Lcsa}{\mathcal{L}_{\mathrm{CSA}}}
\newcommand{\Lrec}{\mathcal{L}_{\mathrm{rec}}}
\newcommand{\Ldiff}{\mathcal{L}_{\mathrm{diff}}}
\newcommand{\vpred}{v_{\mathrm{pred}}}

\newcommand{\norm}[1]{\left\lVert #1\right\rVert}

\newcommand{\ph}{\varphi}
\newcommand{\pdata}{p_{\mathrm{data}}}
\newcommand{\pz}{p_{z}}

\title{Improving the Diffusability of Motion Tokenizer}

\author{%
  Guanhe Huang$^{1}$\thanks{Equal contribution.} \quad
  Songqiao Han$^{2}$\footnotemark[1] \quad
  Tangzheng Lian$^{1}$ \quad
  Oya Celiktutan$^{1}$ \\
  $^{1}$Department of Engineering,
  $^{2}$Department of Mathematics \\
  King's College London, London, United Kingdom \\
  {\small\texttt{\{guanhe.huang,songqiao.han,lian.tangzheng,oya.celiktutan\}@kcl.ac.uk}}
}

\begin{document}

\maketitle
\setcounter{footnote}{0}

\begin{figure}[htbp]
    \centering
    \begin{subfigure}[b]{0.48\textwidth}
        \centering
        \includegraphics[width=\textwidth]{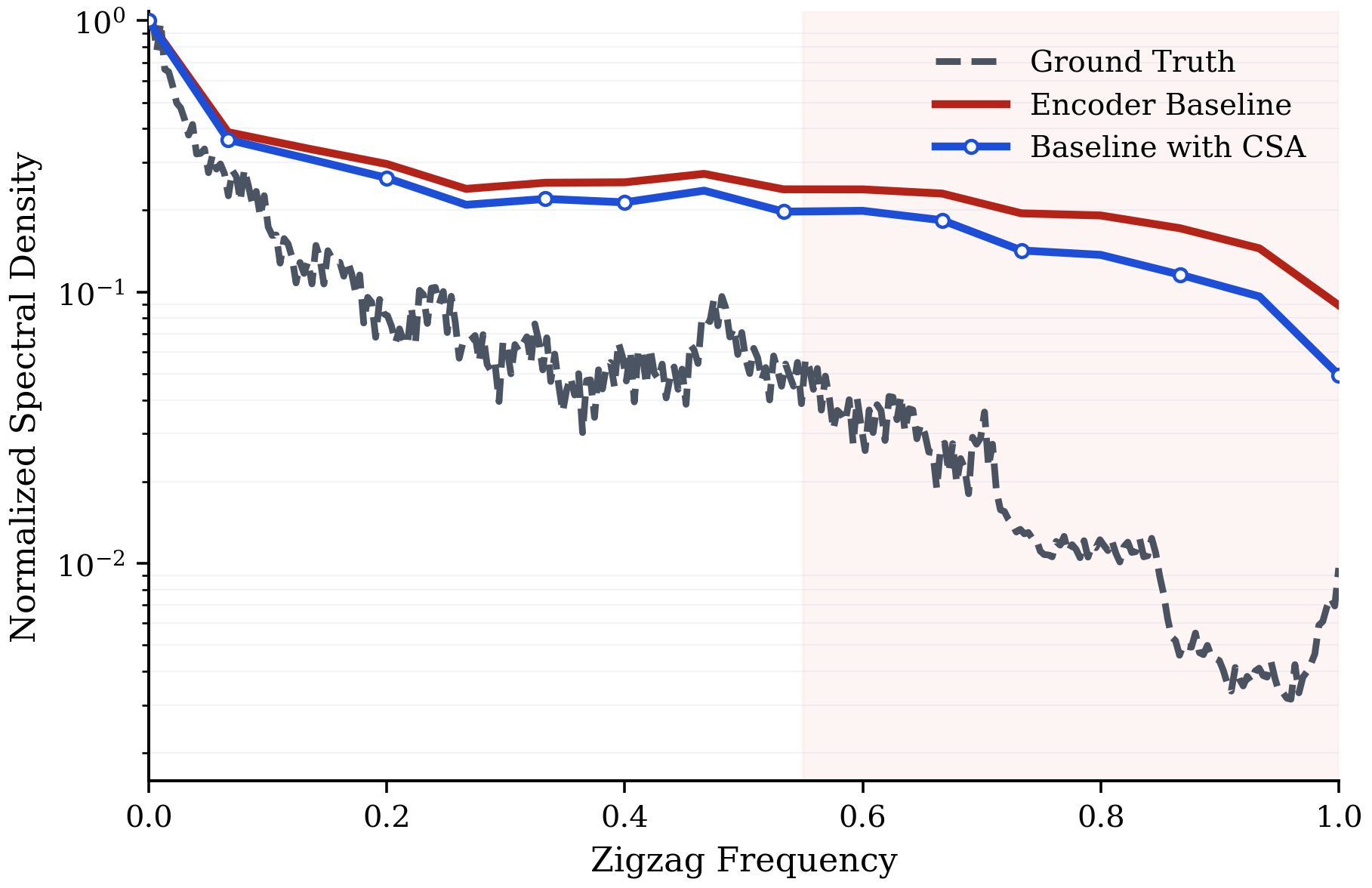}
        \caption{Latent spectral profile}
        \label{fig1a}
    \end{subfigure}
    \hfill
    \begin{subfigure}[b]{0.48\textwidth}
        \centering
        \includegraphics[width=\textwidth]{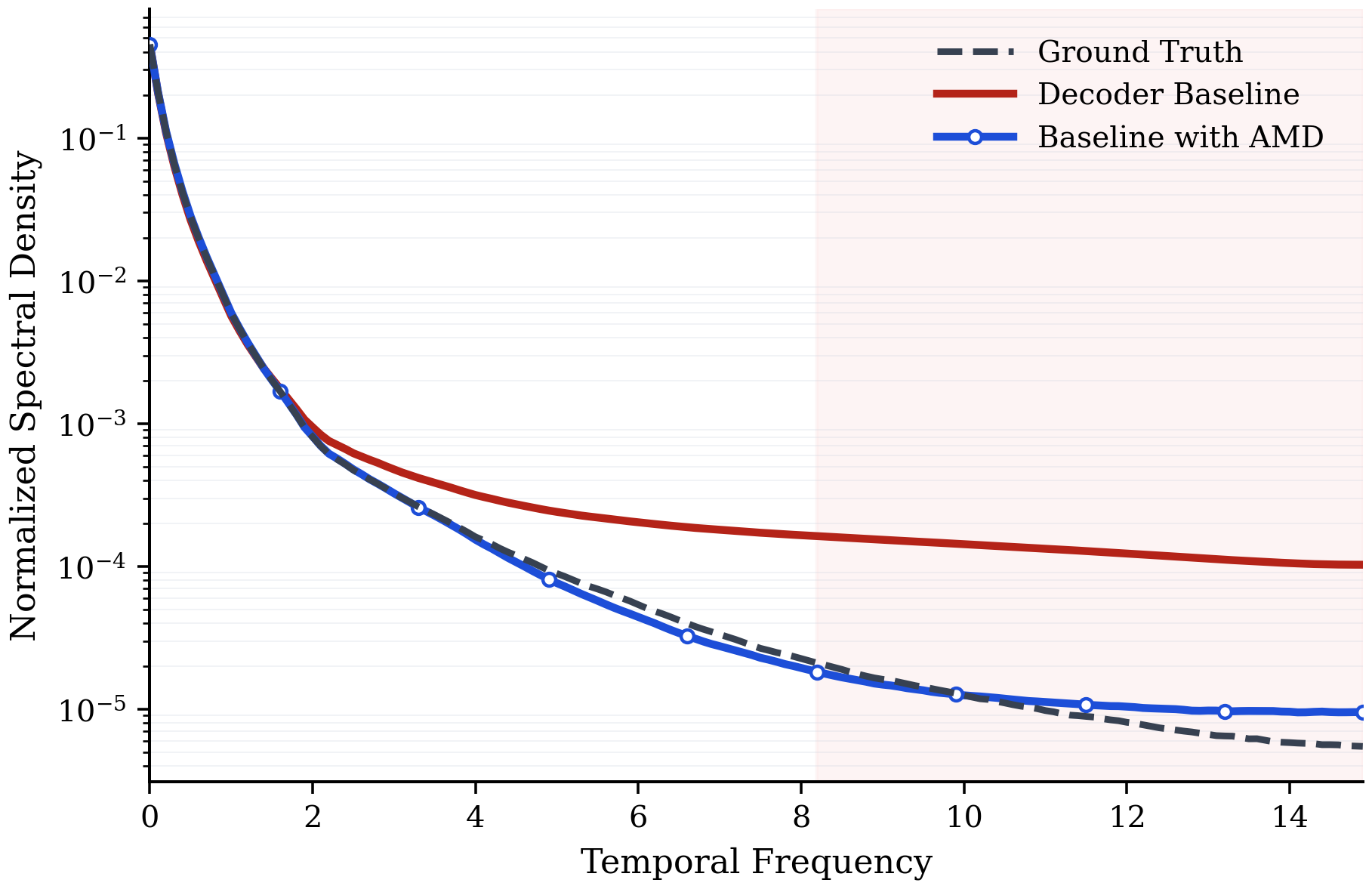}
        \caption{Motion spectral profile}
        \label{fig1b}
    \end{subfigure}
    \caption{
\textbf{Spectral mismatch in motion tokenizers.}
(a) The baseline encoder produces high-frequency shortcut outliers into the latent space. 
Our Critic Spectral Adaptation (CSA) uses the pre-trained diffusion as a frequency-specific critic to suppress this shortcut high-frequency energy.
%
(b) The baseline deterministic decoder introduces severe high-frequency artifacts into the reconstructed motion.
The Alpha-Flow Motion Decoder (AMD) calibrates the spectral profile to closely approximate the real motion dynamics.
}
    \label{fig1}
\end{figure}

\begin{abstract}
Latent diffusion models paired with a variational auto-encoder (VAE) tokenizer have shown promising performance for efficient text-driven human motion generation.
However, motion latent diffusion suffers from a generation-reconstruction trade-off: simply increasing tokenizer capacity enhances reconstruction but fails to proportionally improve text-conditioned generation quality.
We assume this trade-off stems from a spectral mismatch inside the VAE tokenizer: the encoder yields a latent space biased toward high frequencies, and the decoder introduces temporal artifacts.
In this paper, we propose the Diffusable Motion Tokenizer (DiMoT) to improve the \textit{diffusability}~\footnote{Diffusability describes how easily the latent space can be fitted by diffusion.} of the tokenizer, rather than scaling up the diffusion backbone.
To improve text-conditioned optimization, Critic Spectral Adaptation (CSA) leverages a pretrained diffusion model as the critic to suppress high-frequency latent energy.
To preserve motion dynamics, the Alpha-Flow Motion Decoder (AMD) performs generative decoding via average-velocity prediction, enabling efficient one-step decoding.
Extensive experiments demonstrate that DiMoT achieves state-of-the-art performance across various challenging text-to-motion benchmarks, yielding superior motion fidelity and text-motion alignment.
Project page: \url{https://githubhgh.github.io/dimot/}.
\end{abstract}

\section{Introduction}

Generative modeling is entering a \textit{Token Era}, with text~\cite{devlin2019bert,raffel2020exploring,touvron2023llama} and video~\cite{wan2025wan,yu2023language,kondratyuk2023videopoet} tokenization becoming standard practice in large generalist models.
Conversely, motion sequences present unique challenges due to their inherent kinematic structure and strict temporal sensitivity, which are also more expensive and time-consuming to collect.
For example, human joints have strict rotation limits, and any minor temporal misalignment could result in obvious artifacts such as jitter and foot sliding.
Consequently, developing a robust motion tokenizer under these constraints is critical for advancing downstream applications in humanoid robotics~\cite{huang2026learning,fu2024humanplus,he2024omnih2o,he2025asap} and virtual reality~\cite{goel2023humans,jiang2022avatarposer,du2023avatars,li2023ego}.

Latent Diffusion Models (LDMs)~\cite{blattmann2023stable, rombach2022high,saharia2022photorealistic,ramesh2022hierarchical,podell2023sdxl,peebles2023scalable} paired with a pre-trained Variational Autoencoder (VAE) tokenizer have advanced rapidly in motion generation, extending from single-person~\cite{chen2023executing, dai2024motionlcm, jiang2025motionpcm, jin2024act, ji2023c2g2} to multi-person~\cite{geng2026disentangled, huang2026unified, li2024two} scenarios.
Compared with the diffusion backbone, the tokenizer has been less explored, yet it is equally important for high-quality motion generation.
While continuous representations are naturally suited for producing smooth motion~\cite{cho2025discord}, they struggle to capture the discrete nature of conditioning signals (\eg, text).
Furthermore, the tokenizer and generative backbone are typically trained in a decoupled manner, where existing literature often relies on the VAE's reconstruction accuracy as a proxy for generation quality.
However, such an optimal latent space for reconstruction struggles to achieve the same superiority in conditional diffusion generation~\cite{yao2025reconstruction}.
%
%
Previous studies~\cite{skorokhodov2025improving, liu2025delving} attribute this gap to the spectral mismatch: the latent space contains abundant high-frequency energy, whereas raw motion is predominantly low-frequency, which degrades downstream generation.
In this paper, we propose the Diffusable Motion Tokenizer (DiMoT). Specifically, DiMoT couples Critic Spectral Adaptation (CSA) to regularize the encoder's latent space with Alpha-Flow Motion Decoder (AMD) to efficiently preserve motion dynamics.

Firstly, it is well-known that VAEs tend to over-smooth motion sequences into low-frequency content~\cite{kingma2013auto}.
Surprisingly, we empirically find in Fig.~\ref{fig1a} that the encoder yields a latent space biased toward high frequencies absent from the raw motion.
We interpret this as an ill-posed shortcut code in the VAE that easily amplifies back into motion space, while the diffusion backbone struggles to model the text-conditioned distribution with such artificial outliers~\citep{skorokhodov2025improving}.
However, naively applying a uniform high-frequency penalty also suppresses the necessary dynamics for motion reconstruction.
%
We therefore introduce CSA to use the pretrained diffusion model as a text-conditioned critic regularizer~\cite{wang2024larp}.
Moreover, CSA re-weights the regularization loss mode by mode (\ie, Fourier) for spectral reorganization, which adaptively suppresses the VAE's shortcut outliers.
%
By reweighting the latent space critic regularization with the spectral profile,
CSA improves the diffusability of the tokenizer while maintaining comparable reconstruction performance.


Beyond the latent representation, the VAE decoder sets the performance ceiling for generation quality; decoupled training isolates it from the diffusion prior.
%
Consequently, this mismatch implies that a well-optimized latent space does not guarantee accurate motion reconstruction during inference.
%
Previous works~\cite{gu2026bridging, cho2025discord} employ iterative refinement within diffusion decoding, which uses the latent as conditioning signals to directly guide motion generation.
However, the iterative nature of diffusion decoders introduces significant computational overhead, contradicting the primary efficiency of LDMs.
To address it, we introduce the AMD, an efficient one-step generative decoder that learns the average velocity field~\cite{geng2025improved, geng2025mean}. 
%
%
%
As illustrated in Fig.~\ref{fig1b}, AMD's generative decoding matches the real motion frequency distribution, whereas deterministic decoding inherently introduces high-frequency artifacts.
Therefore, the AMD improves the diffusability by calibrating the spectral profile to efficiently preserve the essential motion dynamics.
%

In summary, our core contributions are:
 \begin{itemize}[itemsep=0pt, topsep=3pt, leftmargin=15pt]
\item We introduce \textbf{Critic Spectral Adaptation} (\textbf{CSA}), leveraging a pre-trained LDM as a critic for tokenizer encoder regularization. Guided by the text-conditioned distribution, CSA adaptively suppresses shortcut high-frequency energy via spectral reweighting.
Furthermore, we provide a theoretical analysis detailing how CSA improves generalized latent diffusion frameworks.
\item We propose the \textbf{Alpha-Flow Motion Decoder (AMD)} to replace the tokenizer decoder with the one-step generative decoder. By learning the latent-conditioned average velocity field, AMD calibrates essential motion dynamics without iterative refinement.
\item We integrate these modules into the \textbf{Diffusable Motion Tokenizer (DiMoT)} to conduct comprehensive evaluations across multi-person and single-person benchmarks.
Extensive results show DiMoT demonstrates superior fidelity and text following ability over existing LDM frameworks.
\end{itemize}

\section{Related Works}

\paragraph{Tokenizer Diffusability Bottleneck.} 
%
%
Recent works~\citep{skorokhodov2025improving,leng2025repa,chen2025diffusion,ning2026spectrum} have identified the spectral inconsistency between the latent space and raw data as a primary bottleneck for LDM generation quality.
Specifically, vanilla VAE encoders tend to push the latent space into biased high-frequency modes, which diffusion models struggle to fit~\citep{skorokhodov2025improving}.
Existing solutions attempt to mitigate this by penalizing latent DCT amplitudes~\citep{skorokhodov2025improving}, matching input and latent power spectra~\citep{ning2026spectrum}, aligning latents with external features~\citep{leng2025repa,yao2025reconstruction}, or jointly training the encoder with a diffusion decoder~\citep{chen2025diffusion,zhao2024epsilon}. 
Other approaches regularize latents toward lower frequencies~\citep{liu2025delving} or apply a frequency curriculum during diffusion training without altering the autoencoder~\citep{lai2025toward}. 
In contrast, CSA leverages the same pre-trained diffusion as a per-frequency critic to fine-tune the encoder. This requires no extra module, external feature extractor, or joint training.

\paragraph{Motion Generation in the Latent Space.} 
LDMs~\cite{rombach2022high} are widely adopted in text-to-motion generation due to their effective trade-off between computational efficiency and generation quality~\cite{chen2023executing, dai2024motionlcm, tevet2024closd, geng2026disentangled, li2024two, cho2025discord, gu2026bridging, tuautoregressive, ye2026remogen, pan2025tokenhsi, huang2026unified}.
Early works focus on single-person generation~\citep{chen2023executing}, which are further improved by consistency distillation for real-time inference~\citep{dai2024motionlcm}. 
Different from individual motion generation, multi-person generation is more challenging due to the extra relative orientation and position modeling.
~\citet{geng2026disentangled} applies a decoupled latent space via disentangling the interaction context and individual motion.
Recently, the focus has shifted to unified latent spaces capable of handling variable agents and tasks via shared policies~\citep{pan2025tokenhsi} or domain-specific motion priors~\citep{ye2026remogen}. 
However, coherence between the tokenizer and the diffusion backbone remains underexplored.
Concurrent works bridge this coherence gap via Gaussian mixture regularization~\citep{tuautoregressive} or generative discrete decoding~\citep{cho2025discord}.
These methods either ignore the inherent high-frequency outliers in the decoder output or introduce significant computational overhead.
Instead, we replace the VAE decoder with AMD to preserve the essential motion dynamics, achieving high-fidelity, one-step generative decoding~\cite{zhao2024epsilon}.

\section{Methodology}

\paragraph{Problem Statement.} Given a user-input text $y$, the goal is to generate a natural and smooth motion sequence $x$ that aligns with the language semantics. This can be modeled with a text-conditioned Latent Diffusion Model (LDM), where motion is generated in a latent space and decoded into the final motion sequence.
Standard LDMs use a decoupled approach: training a diffusion model $f_\theta$ strictly on the frozen latents $z = E_\varphi(x)$ of a pre-trained VAE tokenizer ($E_\varphi, D_\psi$).
Rather than scaling up the diffusion backbone, we target improving the tokenizer ($E_\varphi$ and $D_\psi$) to maximize its \textit{diffusability}.

\begin{figure}[t] 
\centering    
\includegraphics[width=1\textwidth]{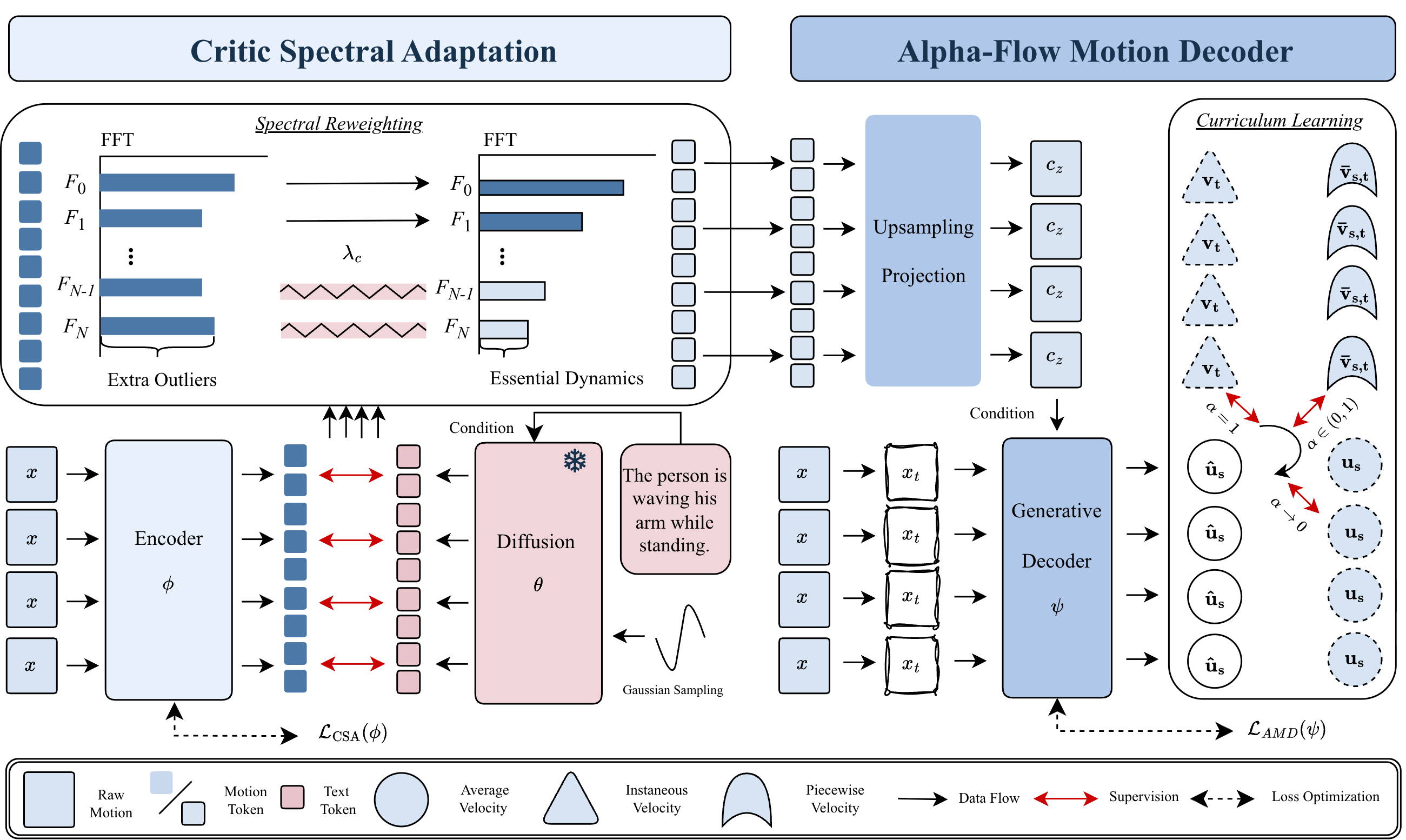}
\caption{
Overview of our DiMoT framework.
Left: Critic Spectral Adaptation (CSA).
CSA utilizes a pre-trained latent diffusion model as a text-conditioned critic, which adaptively suppresses shortcut high-frequency outliers via spectral reweighting. 
Right: Alpha-Flow Motion Decoder (AMD).
By using curriculum learning over trajectory velocity fields in different time scales, AMD gradually learns one-step average-velocity generative decoding with latent conditions.
}
\label{fig2}
\vspace{-3mm}
\end{figure}

To achieve this, we present the Diffusable Motion Tokenizer (DiMoT), a framework that addresses the spectral mismatch and bridges the performance gap between tokenizer reconstruction and diffusion generation.
As illustrated in Fig.~\ref{fig2}, DiMoT consists of two core components: (1) Critic Spectral Adaptation (CSA), which uses a pre-trained diffusion as the critic to adaptively regularize the latent spectral profile of encoder $E_\varphi$, and (2) Alpha-Flow Motion Decoder (AMD), which replaces the decoder $D_\psi$ with one-step generative decoding to preserve temporal motion dynamics.
In this way, DiMoT achieves high-quality text-conditioned generation while simultaneously preserving reconstruction fidelity.

\subsection{Critic Spectral Adaptation (CSA)}\label{sec:spec-critic}

We introduce Critic Spectral Adaptation (CSA) to fine-tune the encoder $E_\varphi$ via spectral reweighting, leveraging a pre-trained text-conditioned diffusion model as a critic~\cite{wang2024larp,poole2022dreamfusion}.
%
%
%
We begin by formulating the spectral critic loss, followed by a theoretical analysis detailing how CSA facilitates the optimization of the downstream diffusion objective.



\paragraph{Diffusion as a Critic.}
By employing the pre-trained diffusion model $f_{\theta_0}$ as a generative critic, CSA optimizes the latent space manifold toward improved text-conditioned generability, rather than trivially fitting the reconstruction objective.
Let $E_\ph$ be the pre-trained motion VAE encoder with the decoder $D_{\psi_0}$ pinned at its baseline parameters $\psi_0$, and let $\mathcal{L}_{\mathrm{rec}}(\ph)$ denote the reconstruction loss as a function of the encoder. 
The frozen text-conditioned rectified-flow latent diffusion model $f_{\theta_0}(x_t, t, y)$ is trained on the baseline tokenizer's latents~\citep{rombach2022high}. 
For motion-prompt pair $(x, y) \sim p_{\mathrm{data}}$ and latent $z = E_\ph(x)$, sample interpolation time $t\!\sim\!\mathcal{U}[0,1]$ and Gaussian noise $\epsilon\!\sim\!\mathcal{N}(0, I)$. Set $x_t = (1{-}t)\,\mathrm{sg}(z) + t\epsilon$. 
The prediction loss/residual is
\begin{equation}\label{eq:residual}
  r(z) \;=\; \mathrm{sg}\!\bigl(f_{\theta_0}(x_t, t, y)\bigr) - (\epsilon - z),
\end{equation}
where $\mathrm{sg}$ is the stop-gradient. 
The stop-gradients keep $f_{\theta_0}$ a fixed critic rather than a trainable model to differentiate through, restricting parameter updates solely to the encoder.
%
%
Finally, the expected squared residual gives the (plain) critic loss $\mathcal{L}_{\mathrm{crit}}(\ph) = \mathbb{E}_{(x,y),t,\epsilon}\,\|r(z)\|^{2}$ (See Fig.~\ref{fig:csa_bar_a} results). 

\paragraph{Spectral Reweighting.}
Inspired by~\citep{skorokhodov2025improving}, we further improve the critic loss in the frequency domain. 
%
The Fast Fourier Transform (FFT) $\mathcal{F}$ decomposes $r(z)$ along the time axis into Fourier modes indexed by $k$, where $k=0$ is the static component and larger $k$ corresponds to faster temporal variations.
The residual amplitude $|\mathcal{F}[r(z)](k)|$ is small at modes the diffusion handles well, and large at the shortcut high-frequency modes the encoder introduces (Fig.~\ref{fig1a}).
However, $\mathcal{L}_{\mathrm{crit}}$ weights every mode equally, so we reweight it across modes to concentrate the encoder's response on the shortcut band:
\begin{equation}\label{eq:Lspec-main}
  \mathcal{L}_{\mathrm{CSA}}(\ph)
  \;=\;
  \mathbb{E}_{(x,y),\,t,\,\epsilon}\!\left[\,
    \frac{1}{N_t}\sum_{k} \tilde w(k)\,\bigl|\mathcal{F}[r(z)](k)\bigr|^{2}
  \right].
\end{equation}
Here, the weight $\tilde w(k) = w(k)/\bar w$ with $w(k) = 1 + \kappa\,d(k)^{p}$ is a static spectral weight ($d(k)\in[0,1]$ denotes FFT distance from $k=0$, $p$ denotes polynomial exponent). The strength knob $\kappa$ controls how aggressively the encoder pressure is biased toward the high-frequency band. 
The gradient on $z$ at each mode is proportional to $\tilde w(k)\,\hat r(k)$ (\Cref{thm:spectral-grad}, Appendix~\ref{app:struct}), so modes already handled by the diffusion receive little encoder gradient.
Algorithm~\ref{alg:crest} summarizes the resulting training step.
\begin{algorithm}[h]
\caption{CSA Training Loop.}\label{alg:crest}
\small
\begin{algorithmic}[1]
\Require encoder $E_\ph$, frozen diffusion $f_{\theta_0}$, FFT length $N_t$, 
total steps $S$
\State Pre-compute spectral weights $\tilde w(k) = (1 + \kappa\,d(k)^p)/\bar w$ for $k = 0,\ldots,N_t{-}1$ \Comment{static, FFT-geometry only}
\For{step $i = 1,\ldots,S$}
  \State Sample $(x, y) \sim p_{\mathrm{data}}$,\quad $t \sim \mathcal{U}[0,1]$,\quad $\epsilon \sim \mathcal{N}(0,I)$ \Comment{joint motion-prompt batch}
  \State $z \gets E_\ph(x)$,\quad $x_t \gets (1{-}t)\,\mathrm{sg}(z) + t\epsilon$ \Comment{noised interpolant, encoder takes $x$ only}
  \State $r \gets \mathrm{sg}\!\bigl(f_{\theta_0}(x_t, t, y)\bigr) - (\epsilon - z)$ \Comment{Eq.~\eqref{eq:residual}, text $y$ enters here}
  \State $\mathcal{L}_{\mathrm{CSA}} \gets N_t^{-1}\sum_k \tilde w(k)\,|\mathcal{F}[r](k)|^{2}$ \Comment{Eq.~\eqref{eq:Lspec-main}, FFT along temporal axis}
  \State Update $\ph$ on $\mathcal{L}_{\mathrm{rec}}(\ph) + \lambda_c\min(1, i/W)\,\mathcal{L}_{\mathrm{CSA}}(\ph)$ \Comment{$\psi_0, \theta_0$ frozen, $W$-step warmup}
\EndFor
\end{algorithmic}
\end{algorithm}


\paragraph{CSA Improves Downstream Diffusion.}\label{sec:csa-text}
While $E_\varphi$ ($\varphi$ denotes the encoder parameters) operates solely on motion inputs, CSA guides the encoder's gradients by incorporating the prompt $y$ via the critic's text-conditioned prediction $\mathrm{sg}(f_{\theta_0}(x_t,t,y))$. 
Moreover, we provide the theoretical foundations detailing how it improves general conditional diffusion optimization within this learned latent space.
%

Define the text-conditioned diffusion training loss as 
$\mathcal{L}_{\mathrm{diff}}(\theta;\ph) \coloneqq \mathbb{E}_{(x,y),t,\epsilon}\,\|f_\theta(x_t,t,y)-(\epsilon-z)\|^{2}$. 
Under reasonable conditions, we analyze the CSA regularization from the bilevel optimization theory~\cite{lorraine2020optimizing,sener2018multi,poole2022dreamfusion}. 
%
Consequently, we prove that CSA translates into a strict improvement in $\mathcal{L}_{\mathrm{diff}}$ as follows:

\begin{theorem}[\textit{Diffusion-loss improvement}]
\label{thm:diff-loss}
Adopt the assumptions and notation of Theorem~\ref{thm:info-preservation} (Appendix~\ref{app:pareto}).  Suppose also $\theta_0 = \arg\min_\theta \mathcal{L}_{\mathrm{diff}}(\theta; \ph_0)$ and $g_{\mathrm{crit}}^\top H^{-1} g > 0$, where $g_{\mathrm{crit}}\coloneqq\nabla_\ph\mathcal{L}_{\mathrm{diff}}(\theta_0;\ph_0)$ and $H\coloneqq\nabla_\ph^2\mathcal{L}_{\mathrm{rec}}(\ph_0)$. Then for all sufficiently small $\lambda_c > 0$,
\begin{equation}\label{eq:diff-improvement}
  \min_\theta \mathcal{L}_{\mathrm{diff}}\!\bigl(\theta;\,\ph^{*}(\lambda_c)\bigr)
  \;<\;
  \min_\theta \mathcal{L}_{\mathrm{diff}}\!\bigl(\theta;\,\ph_0\bigr).
\end{equation}
The alignment condition is treated as a working assumption (Appendix~\ref{app:diff-loss}).
\end{theorem}

Intuitively, at the CSA-trained encoder $\ph^{*}(\lambda_c)$, training the diffusion from scratch yields \textbf{a strictly lower training loss} than at the baseline $\ph_0$.
Meanwhile, when fine-tuning, reconstruction is theoretically preserved at first order in $\lambda_c$ along the same trajectory (\Cref{thm:info-preservation}, Appendix~\ref{app:pareto}). 
%
Even without optimality of the baseline diffusion, \Cref{cor:finetune} yields the same conclusion for fixed $\theta_0$.

As a cross-modality check, we apply CSA to a class-conditional \textbf{ImageNet}~\cite{deng2009imagenet,imagenet15russakovsky} pipeline (\Cref{app:imagenet-csa}). In line with \Cref{thm:diff-loss}, generation FID improves at the evaluated diffusion training checkpoints (i.e., $100k$, $200k$ and $300k$ steps). Reconstruction FID remains close to baseline, suggesting information is reorganized rather than removed.


\subsection{Alpha-Flow Motion Decoder}\label{sec:Alpha-Flow}



As shown in Fig.~\ref{fig1}, while deterministic decoding inherently produces this high-frequency noise, generative decoding can match the frequency distribution of real motion.
Compared to the VAE encoder, the VAE decoder directly determines the upper bound of LDM generation performance, while it only plays the static role as a passive receiver in a mismatched way~\cite{zhao2024epsilon}.
Such a mismatch becomes increasingly disruptive for the motion tokenizer.
Iterative generative decoding can mitigate it, which also introduces prohibitive computational overhead.
To this end, we propose the Alpha-Flow Motion Decoder (\textbf{AMD}) in Fig.~\ref{fig2}, a latent-conditioned one-step generative decoder that preserves the efficiency while restoring motion details lost by standard VAE decoding~\cite{skorokhodov2025improving, liu2025delving}.

\paragraph{Latent Condition Upsampling Projection.}
Given latent tokens $z \in \mathbb{R}^{N_z \times d_z}$ from a frozen diffusion backbone, we aim to recover a frame-aligned condition sequence $c_z \in \mathbb{R}^{T \times d_C}$.
%
%
Unlike~\cite{cho2025discord}, our continuous latent tokens avoid strict frame-to-frame correspondence, where each token is aligned with a coarse temporal segment.
Rather than naively interpolating these tokens, we map the compressed memory to target frames using a Transformer decoder for upsampling projection.
We first linearly project each latent token into the conditioning space, 
$
m = \mathrm{Proj}(z) \in \mathbb{R}^{N_z \times d_C}
$, 
which serves as the memory sequence.
Meanwhile, we construct a set of learnable frame queries
$
q \in \mathbb{R}^{T \times d_C},
$
where each query represents one target frame position.
After adding positional encoding to both the memory and query sequences, we feed them into a transformer decoder, where the queries interact with each other through self-attention and retrieve motion information from the latent memory through cross-attention:
$
c_z = \mathrm{TransDec}(q, m) \in \mathbb{R}^{T \times d_C}.
$
This global cross-attention ensures each condition vector $c_z$ captures the full semantic context, providing the fine-grained temporal alignment required for accurate motion generation.


\paragraph{Generative Decoding on Average Velocity Field.}
After obtaining the latent condition, we apply AMD~\cite{zhang2025alphaflow} as a generative backbone to generate the motion sequences.
Compared to vanilla rectified flow matching~\cite{esser2024scaling}, AMD learns to predict the average velocity field in a one-step setting, without relying on costly iterative refinement.
In detail, vanilla flow matching learns local denoising directions instantaneously, whereas AMD predicts a long-range transport toward motion target $X$ via velocity integration~\cite{geng2025mean}.
Moreover, AMD applies curriculum training to achieve smooth transitions between different trajectory flow matching, which shows better convergence and stability.

Following MeanFlow~\citep{geng2025mean}, we introduce the average velocity field for one-step diffusion decoding.
Given a destination time $r_{d} < t$, the decoder $u_{\psi}$ predicts an average transition velocity $u(x_t, r_d, t, z)$ over the interval $[r_d,t]$ conditioned on latent space $z$ as follows:
\begin{equation}\label{eq:meanflow-avg}
u(x_t, r_d, t, z)
=
\frac{1}{t-r_d}\int_{r_d}^t v(x_{\tau}, \tau, z)\, d\tau,
\end{equation}
where $x$ is the clean motion sample and Gaussian noise $\epsilon \sim \mathcal{N}(0,I)$,
$
x_t = (1-t)x + t\epsilon, v_t = \epsilon - x,
$
where $t \in [0,1]$, $x_0=x$, and $x_1=\epsilon$.
By conditioning explicitly on the destination time, the model can represent large trajectory jumps that are necessary for direct one-step decoding based on \textit{MeanFlow identity}.
However, the original MeanFlow's training target is not tractable and depends on the network prediction. By differentiating the integral definition (see Appendix~\ref{app:meanflow-identity} for derivation), the average velocity derivative target is calculated via the Jacobian-vector product of the decoder network:
\begin{equation}\label{eq:meanflow-jvp}
u(x_t, r_d, t, z)
= v(x_t, t, z) - (t - r_d) \frac{d}{dt} u_{\psi-}(x_t, r_d, t, z).
\end{equation}
%

In practice, MeanFlow training configuration sets $t=r_d$ for $75\%$ of samples, providing instantaneous flow-matching supervision~\cite{geng2025mean}.
AMD instead uses $t=r_d$ for $50\%$ of each batch, while the remaining samples with $r_d<t$ learn interval-average velocities for one-step decoding.
The $\alpha$ schedule and training details are provided in Appendix~\ref{imple_app}.


\paragraph{Alpha-Flow Curriculum Learning}
To mitigate such reliance on trivial instantaneous velocity prediction, AMD performs one-step average velocity supervision by gradually shifting from fast-converging trajectory flow matching to intractable MeanFlow optimization.
%
%
In practice, AMD enforces consistency between interval predictions defined over $(t,s,r_d)$:
\begin{equation}
    s = \alpha r_d + (1-\alpha)t, \alpha \in (0,1], \quad x_s = x_t - (t-s)\tilde{v}_{s,t},
\end{equation}
where $\alpha$ is the time-variant scheduler, and $x_s$ is the motion state at timestep $s$.
Based on this connection, we train the decoder with an $\alpha$-curriculum rather than a fixed objective (\eg, $s=t$ for vanilla flow matching).
%
%
The AMD objective is formally defined as:
\begin{equation}
\mathcal{L}_\mathrm{AMD}(\psi)
=
\mathbb{E}_{x,\epsilon,t,r_d}
\left[
\frac{1}{\alpha}
\left\|
u_{\psi}(x_t, r_d, t, z)
-
\Bigl(
\alpha \tilde{v}_{s,t}
+
(1-\alpha)u_{\psi^-}(x_s, r_d, s, z)
\Bigr)
\right\|_2^2
\right],
\end{equation}
where $\tilde{v}_{s,t} = v_t$ is used to form the shift target in the endpoint value.
%
Specifically, when $\alpha=1$, it reduces to trajectory flow matching, which provides stable transition supervision.
When $\alpha \rightarrow 0$, it recovers the MeanFlow regime, which is better aligned with the final goal of one-step generation.

Notably, the latent can come from text-conditioned diffusion backbones or VAE encoders.
A decoder trained with a tokenizer latent may still face a train-test mismatch at inference time, while the diffusion backbone could produce training samples with high variance.
To reduce this mismatch, we use a mixed training scheme by randomly replacing the latent samples produced.
%
%

At inference time, we instantiate decoding with a single interval from $t=1$ to $r_d=0$, starting from Gaussian noise in raw motion space.
Conditioned on the latent token, the decoder predicts the whole average velocity field of raw motion sequences in one step:
$
\hat{x} = x_1 - u_{\psi}(x_1, 0, 1, z).
$
In this way, we use the AMD to replace deterministic decoding with efficient but more effective motion realization, preserving the coherent high-frequency dynamics (See the video in the supplementary material).

\subsection{DiMoT Justification}

\textbf{Training the Tokenizer for Generation}
While DiMoT optimizes the tokenizer, it is not limited to maximizing standard reconstruction accuracy.
Instead, DiMoT enhances motion generation by improving the tokenizer's diffusability. 
It structures the latent space to be more easily modeled by the diffusion process. 
We evaluate this by showing that DiMoT improves text-conditioned generation performance while maintaining comparable reconstruction fidelity. 
To isolate the tokenizer's contribution, we keep the downstream diffusion backbone architecture and training configuration strictly fixed (See Appendix~\ref{imple_app}).

\textbf{Optimal Latent Spectral Profile}
High-frequency energy in the latent space is more challenging to suppress, as latent spaces lack the tractable ground truth for spectral alignment. 
While excessively high-frequency energy degrades diffusion generation, naively eliminating it also compromises sample quality~\cite{skorokhodov2025improving}. 
Consequently, prior methods~\cite{liu2025delving} rely on mathematical transformations (\eg, downsampling) as anchors.
In contrast, CSA provides a flexible anchoring mechanism that adaptively penalizes high-frequency components, exclusively for samples misaligned with the critic distribution.

\section{Experiments}\label{sec:experiments}

\subsection{Datasets}
We utilize InterHuman~\cite{liang2024intergen} and Inter-X~\cite{xu2024inter} datasets for the evaluation of text-conditioned interaction generation performance. 
We also provide HumanML3D~\cite{guo2022generating} results in Tab.~\ref{tab:single_extra}, but this paper focuses on the dual-person task. This setting imposes stricter constraints that more effectively expose the temporal artifacts and high-frequency outliers our method addresses.
%

The InterHuman and InterX datasets contain 7,779 and 11,388 interaction sequences, respectively, where each sequence is illustrated with 3 textual annotations.
%
%
InterHuman follows the SMPL~\cite{SMPL:2015} skeleton representation with 22
joints, including the root joint. Each interaction sequence at the $i$-th frame is represented by:
$
    x^{i}=\left[\mathbf{j}_{g}^{p}, \mathbf{j}_{g}^{v}, \mathbf{j}^{r}, \mathbf{c}
    ^{f}\right]
$
which is the collection of joint positions
$\mathbf{j}_{g}^{p}\in \mathbb{R}^{2 \times 22 \times 3}$, joint velocities $\mathbf{j}
_{g}^{v}\in \mathbb{R}^{2 \times 22 \times3}$, 6D representation rotations
$\mathrm{j}^{r}\in \mathbb{R}^{2 \times 21 \times 6}$, and binary foot-ground contact
features $\mathbf{c}^{f}\in \mathbb{R}^{ 2 \times4}$, resulting in a total input
dimension of 524.
Inter-X follows the SMPL-X~\cite{SMPL-X:2019} skeleton representation,
comprising 54 body and hand joints, accompanied by root orientation and translation.
Each interaction sequence could be represented by
$
    x^{i}=\left[\mathbf{r}_{g}^{p}, \mathbf{j}^{r}\right]
$
comprising root joint positions $\mathbf{r}_{g}^{p}\in \mathbb{R}^{2 \times 3}$ and Euler angle representation rotations $\mathrm{j}^{r}\in \mathbb{R}^{2 \times 55 \times 3}$, for a total input dimension of 336.

We employ the evaluation metrics following previous studies~\cite{liang2024intergen,
guo2022generating}. (See Appendix~\ref{metric_app} for more details).
The implementation of CSA, AMD, and diffusion backbone can be found in Appendix~\ref{imple_app}.

\begin{table*}[t]
    \caption{Quantitative evaluation on the InterHuman~\cite{liang2024intergen} and InterX test sets. $\pm$ indicates a 95\% confidence interval (10 runs) and $\rightarrow$ means the closer to ground truth the better. 
    The deep blue background indicates the best, while the light blue background indicates the second-best. '$^{\dag}$' denotes self-reproduced results.
    }
    \label{main_tab}
    \centering
    \resizebox{1\textwidth}{!}{
    \begin{tabular}{cl|ccc|c|c|c|c}
        \toprule 
        \multirow{2}{*}{Dataset} & \multirow{2}{*}{Method} & \multicolumn{3}{c|}{R Precision$\uparrow$} & \multirow{2}{*}{FID$\downarrow$} & \multirow{2}{*}{MM Dist$\downarrow$} & \multirow{2}{*}{Diversity$\rightarrow$} & \multirow{2}{*}{MModality$\uparrow$} \\
        \cmidrule{3-5}                                     
        & & Top 1 & Top 2 & Top 3 & & & & \\
        \midrule \midrule 
        
        \multirow{11}{*}{\begin{tabular}[c]{@{}c@{}}Inter\\Human\end{tabular}} 
        & Ground Truth                      & $0.452^{\pm0.008}$             & $0.610^{\pm0.009}$             & $0.701^{\pm0.008}$             & $0.273^{\pm0.007}$             & $3.755^{\pm0.008}$             & $7.948^{\pm0.064}$             & -                              \\
        \cmidrule{2-9}                                                                                                             
        & TEMOS~\cite{petrovich2022temos}   & $0.224^{\pm0.010}$             & $0.316^{\pm0.013}$             & $0.450^{\pm0.018}$             & $17.375^{\pm0.043}$             & $6.342^{\pm0.015}$             & $6.939^{\pm0.071}$             & $0.535^{\pm0.014}$             \\
        & T2M~\cite{guo2022generating}      & $0.238^{\pm0.012}$             & $0.325^{\pm0.010}$             & $0.464^{\pm0.014}$             & $13.769^{\pm0.072}$             & $5.731^{\pm0.013}$             & $7.046^{\pm0.022}$             & $1.387^{\pm0.076}$             \\
        & MDM~\cite{tevet2022human}         & $0.153^{\pm0.012}$             & $0.260^{\pm0.009}$             & $0.339^{\pm0.012}$             & $9.167^{\pm0.056}$             & $7.125^{\pm0.018}$             & $7.602^{\pm0.045}$             & \cellcolor{blue!30}$2.350^{\pm0.080}$    \\
        & ComMDM~\cite{shafir2023human}     & $0.223^{\pm0.009}$             & $0.334^{\pm0.008}$             & $0.466^{\pm0.010}$             & $7.069^{\pm0.054}$             & $6.212^{\pm0.021}$             & $7.244^{\pm0.038}$             & $1.822^{\pm0.052}$             \\
        & InterGen~\cite{liang2024intergen} & $0.371^{\pm0.010}$             & $0.515^{\pm0.012}$             & $0.624^{\pm0.010}$             & $5.918^{\pm0.079}$             & $5.108^{\pm0.014}$             & $7.387^{\pm0.029}$             & \cellcolor{blue!15}$2.141^{\pm0.063}$ \\
        & TIMotion~\cite{wang2025timotion}  & $0.491^{\pm0.005}$    & \cellcolor{blue!15}$0.648^{\pm0.004}$    & \cellcolor{blue!15}$0.724^{\pm0.004}$    & $5.433^{\pm0.080}$             & $3.775^{\pm0.001}$    & $8.032^{\pm0.030}$             & $0.952^{\pm0.032}$             \\
        & in2IN~\cite{ponce2024in2in}       & $0.425^{\pm0.008}$             & $0.576^{\pm0.008}$             & $0.662^{\pm0.009}$             & $5.535^{\pm0.120}$             & $3.803^{\pm0.002}$             & $7.953^{\pm0.047}$ & $1.215^{\pm0.023}$             \\
        & InterMask~\cite{javed2024intermask} & $0.449^{\pm0.004}$            & $0.599^{\pm0.005}$             & $0.683^{\pm0.004}$             & $5.154^{\pm0.061}$             & $3.790^{\pm0.002}$             & \cellcolor{blue!30}$7.944^{\pm0.033}$    & $1.737^{\pm0.020}$             \\
        & DHVAE~\cite{geng2026disentangled}  & \cellcolor{blue!15}$0.496^{\pm0.004}$             & $0.647^{\pm0.006}$             & $0.720^{\pm0.005}$             & \cellcolor{blue!15}$5.015^{\pm0.085}$             & \cellcolor{blue!30}$3.772^{\pm0.002}$             & \cellcolor{blue!15}$7.952^{\pm0.045}$             & $0.804^{\pm0.030}$          \\
        
        & MLD++$^{\dag}$~\cite{motionlcm-v2}  & $0.412^{\pm0.007}$ & $0.594^{\pm0.003}$ & $0.688^{\pm0.005}$ & $5.046^{\pm0.084}$ & $3.787^{\pm0.001}$ & $8.084^{\pm0.023}$ & $0.613^{\pm0.028}$ \\
        \cmidrule{2-9}
        & DiMoT                             & \cellcolor{blue!30}$0.519^{\pm0.004}$             & \cellcolor{blue!30}$0.663^{\pm0.008}$             & \cellcolor{blue!30}$0.739^{\pm0.006}$             & \cellcolor{blue!30}$4.716^{\pm0.080}$             & \cellcolor{blue!15}$3.775^{\pm0.001}$             & $7.970^{\pm0.035}$             & $0.650^{\pm0.013}$             \\

        \midrule \midrule 
        
        \multirow{11}{*}{\begin{tabular}[c]{@{}c@{}}InterX\end{tabular}} 

       & Ground Truth                      & $0.429^{\pm0.004}$             & $0.626^{\pm0.003}$             & $0.736^{\pm0.003}$             & $0.002^{\pm0.002}$             & $3.536^{\pm0.013}$             & $9.734^{\pm0.078}$             & -                              \\
       \cmidrule{2-9}                                                                                                                                                                    
       & TEMOS~\cite{petrovich2022temos}   & $0.092^{\pm0.003}$             & $0.171^{\pm0.003}$             & $0.238^{\pm0.002}$             & $29.258^{\pm0.069}$             & $6.867^{\pm0.013}$             & $4.738^{\pm0.078}$             & $0.672^{\pm0.041}$             \\
       & T2M~\cite{guo2022generating}      & $0.184^{\pm0.010}$             & $0.298^{\pm0.006}$             & $0.396^{\pm0.005}$             & $5.481^{\pm0.382}$             & $9.576^{\pm0.006}$             & $5.771^{\pm0.151}$             & $2.761^{\pm0.042}$             \\
       & MDM~\cite{tevet2022human}         & $0.203^{\pm0.009}$             & $0.329^{\pm0.007}$             & $0.426^{\pm0.005}$             & $23.701^{\pm0.057}$             & $9.548^{\pm0.014}$             & $5.856^{\pm0.077}$             & \cellcolor{blue!15}$3.490^{\pm0.061}$ \\
       & MDM(GRU)~\cite{tevet2022human}    & $0.179^{\pm0.006}$             & $0.299^{\pm0.005}$             & $0.387^{\pm0.007}$             & $32.617^{\pm0.122}$             & $9.557^{\pm0.019}$             & $7.003^{\pm0.134}$             & $3.430^{\pm0.035}$             \\
       & ComMDM~\cite{shafir2023human}     & $0.090^{\pm0.002}$             & $0.165^{\pm0.004}$             & $0.236^{\pm0.004}$             & $29.266^{\pm0.067}$             & $6.870^{\pm0.017}$             & $4.734^{\pm0.067}$             & $0.771^{\pm0.053}$             \\
       & InterGen~\cite{liang2024intergen} & $0.207^{\pm0.004}$             & $0.335^{\pm0.005}$             & $0.429^{\pm0.005}$             & $5.207^{\pm0.216}$             & $9.580^{\pm0.011}$             & $7.788^{\pm0.208}$             & \cellcolor{blue!30}$3.686^{\pm0.052}$    \\
       & InterMask~\cite{javed2024intermask} & $0.403^{\pm0.005}$ & $0.595^{\pm0.004}$ & $0.705^{\pm0.005}$    & $0.399^{\pm0.013}$             & $3.705^{\pm0.017}$ & \cellcolor{blue!15}$9.046^{\pm0.073}$             & $2.261^{\pm0.081}$             \\                                        
       & DHVAE~\cite{geng2026disentangled}  & \cellcolor{blue!15}$0.442^{\pm0.005}$ & \cellcolor{blue!15}$0.638^{\pm0.005}$ & \cellcolor{blue!15}$0.745^{\pm0.004}$ & \cellcolor{blue!15}$0.339^{\pm0.015}$ & \cellcolor{blue!15}$3.604^{\pm0.020}$ & \cellcolor{blue!30}$9.378^{\pm0.065}$ & $2.505^{\pm0.047}$ \\
        & MLD++$^{\dag}$~\cite{motionlcm-v2} & $0.433^{\pm0.005}$ & $0.625^{\pm0.007}$ & $0.732^{\pm0.006}$ & $0.446^{\pm0.017}$ & $3.659^{\pm0.028}$ & $4.664^{\pm0.062}$ & $1.611^{\pm0.115}$ \\
       \cmidrule{2-9}
       & DiMoT & \cellcolor{blue!30}$0.450^{\pm0.007}$ & \cellcolor{blue!30}$0.642^{\pm0.004}$ & \cellcolor{blue!30}$0.746^{\pm0.004}$ & \cellcolor{blue!30}$0.335^{\pm0.013}$ & \cellcolor{blue!30}$3.562^{\pm0.030}$ & $4.648^{\pm0.060}$ & $2.427^{\pm0.112}$ \\


       \bottomrule
    \end{tabular}
    }

\end{table*}

\begin{figure*}[t]
    \centering
    
    {\setlength{\tabcolsep}{80pt}
    \begin{tabular}{*{8}{c}}
    \toprule
        \multicolumn{4}{c}{\large\bfseries MLD++} & 
        \multicolumn{4}{c}{\large\bfseries DiMoT} \\
        \bottomrule
    \end{tabular}
    }
    
    \vspace{0.2cm}
    {\itshape {The first person sprints towards the second, the second dismisses their request with a hand gesture.}}
    
    {\setlength{\tabcolsep}{0pt}
    \begin{tabular}{cccc @{\hspace{2.5pt}} | @{\hspace{2.5pt}} cccc}
    
        \includegraphics[width=0.12\textwidth]{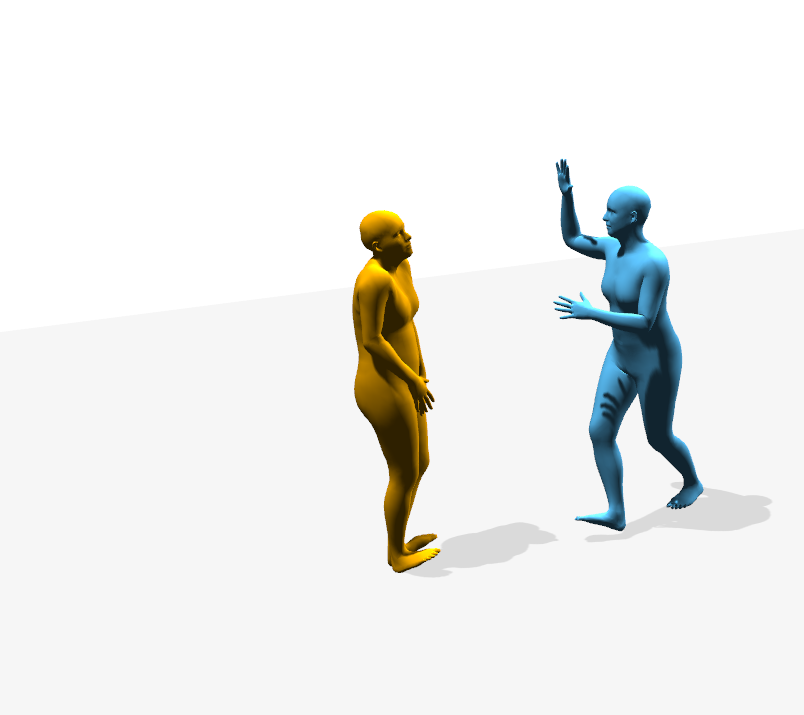} &
        \includegraphics[width=0.12\textwidth]{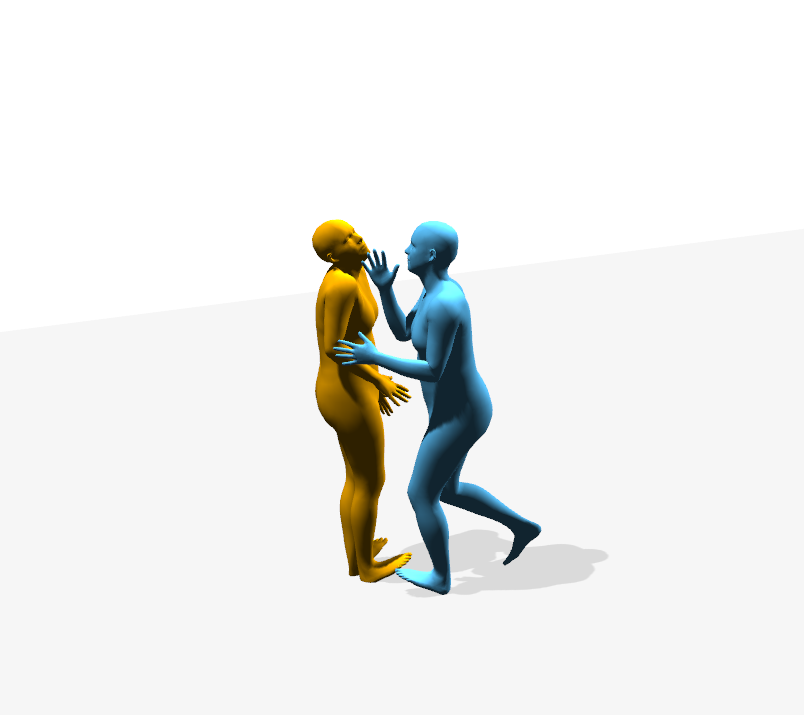} &
        \includegraphics[width=0.12\textwidth]{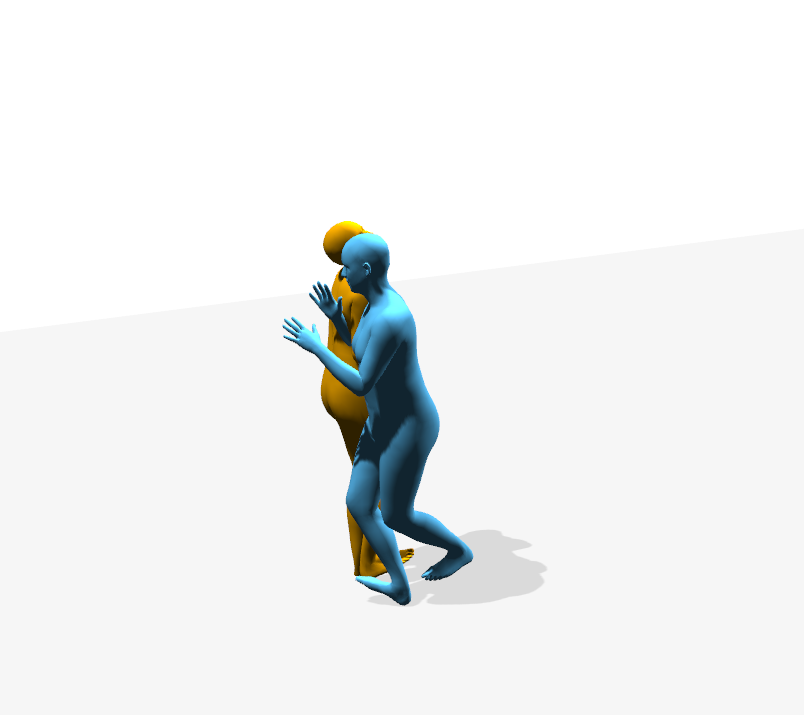} &
        \includegraphics[width=0.12\textwidth]{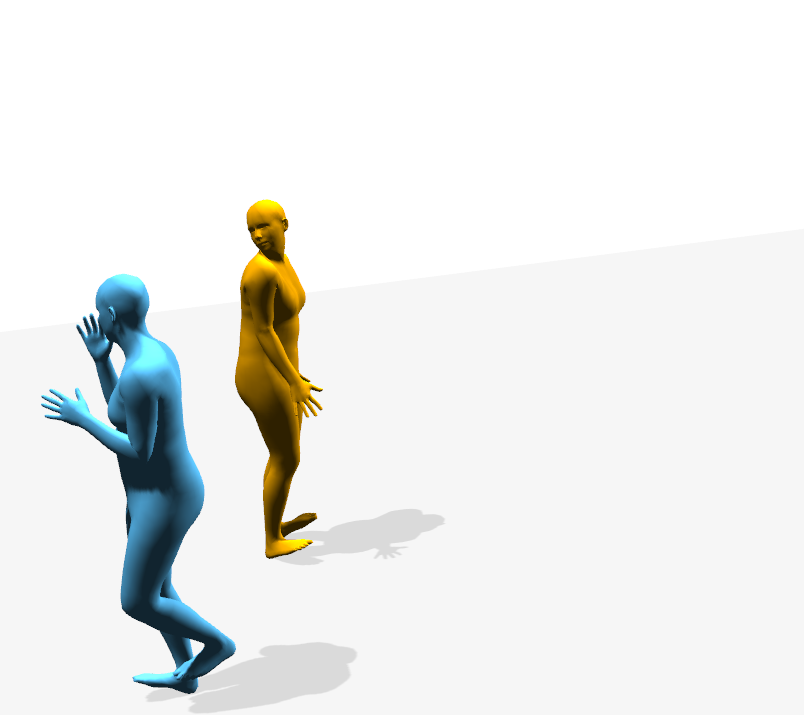} &
        \includegraphics[width=0.12\textwidth]{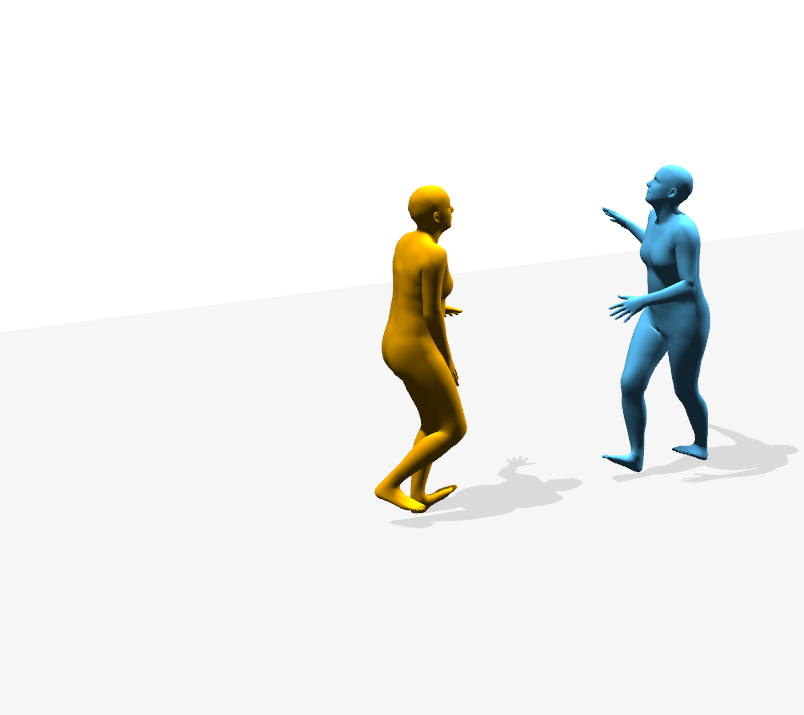} &
        \includegraphics[width=0.12\textwidth]{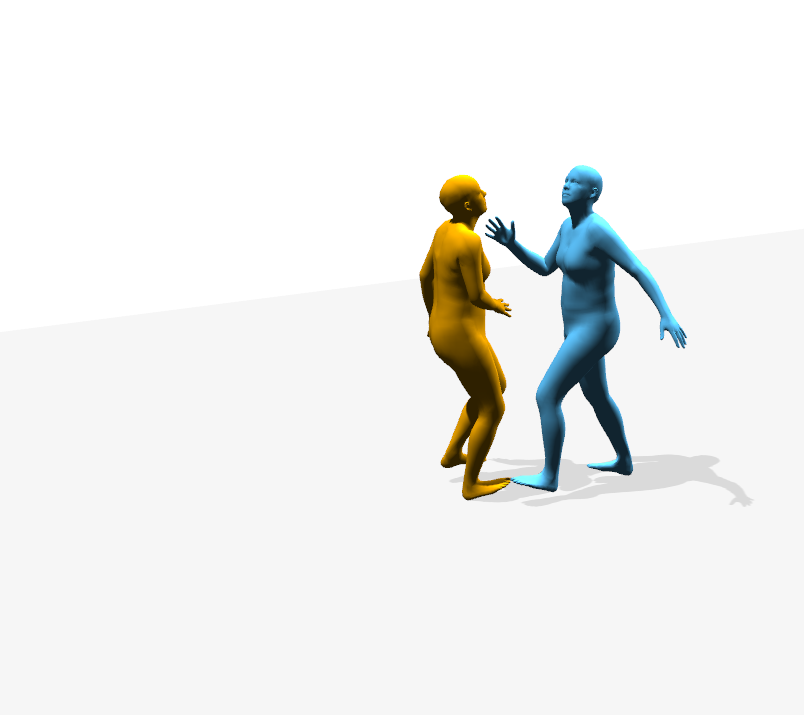} &
        \includegraphics[width=0.12\textwidth]{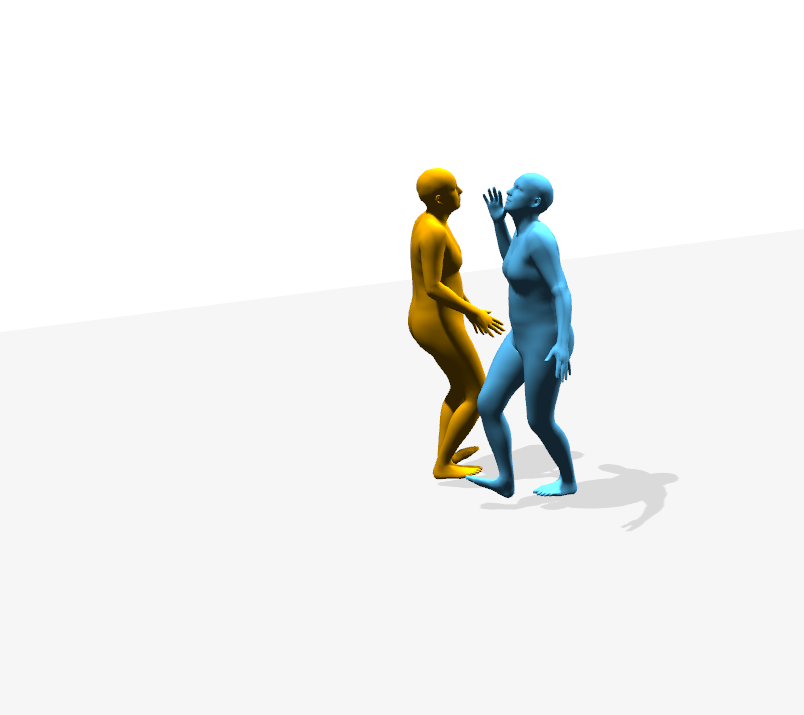} &
        \includegraphics[width=0.12\textwidth]{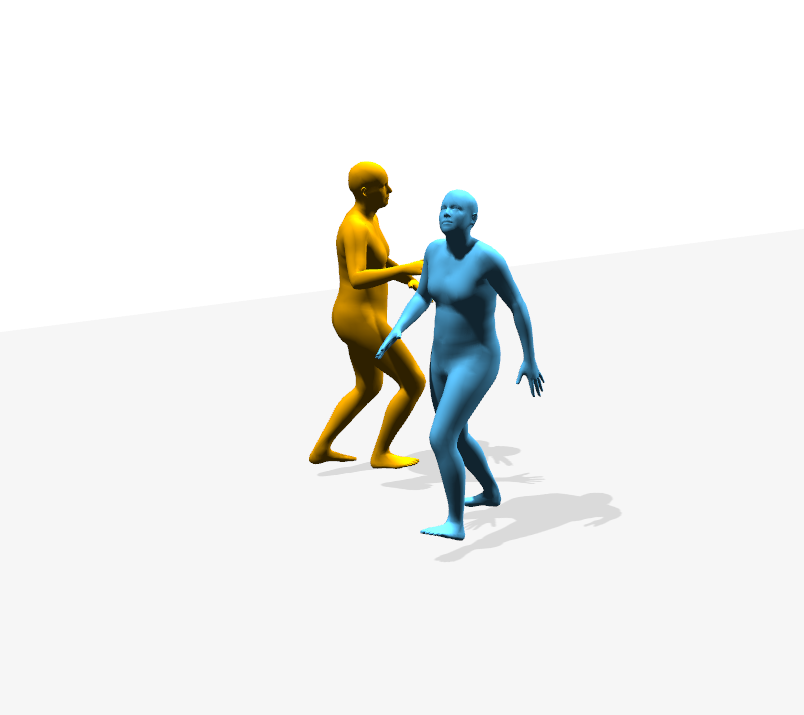}  \\ 
    \end{tabular}}

        {\itshape {They are dancing latin, stomping their feet and swaying their hips.}}
    
    {\setlength{\tabcolsep}{0pt}
    \begin{tabular}{cccc @{\hspace{2.5pt}} | @{\hspace{2.5pt}} cccc}
    
        \includegraphics[width=0.12\textwidth]{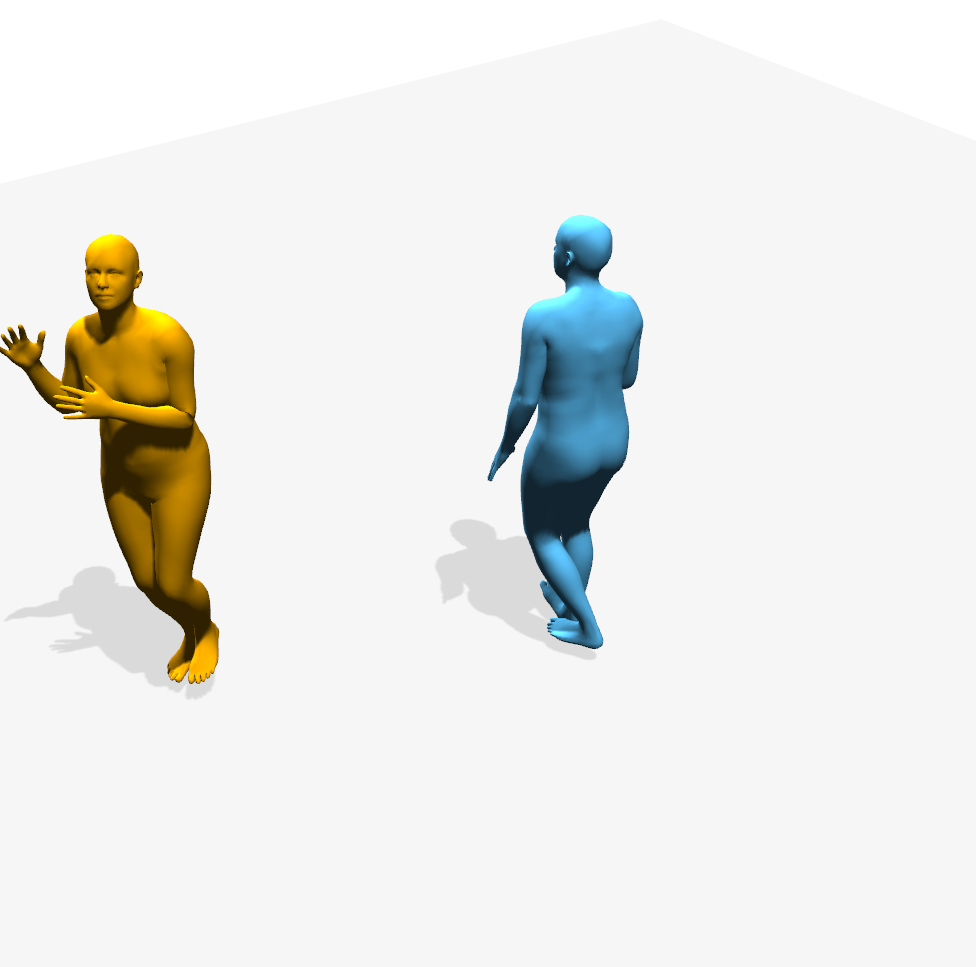} &
        \includegraphics[width=0.12\textwidth]{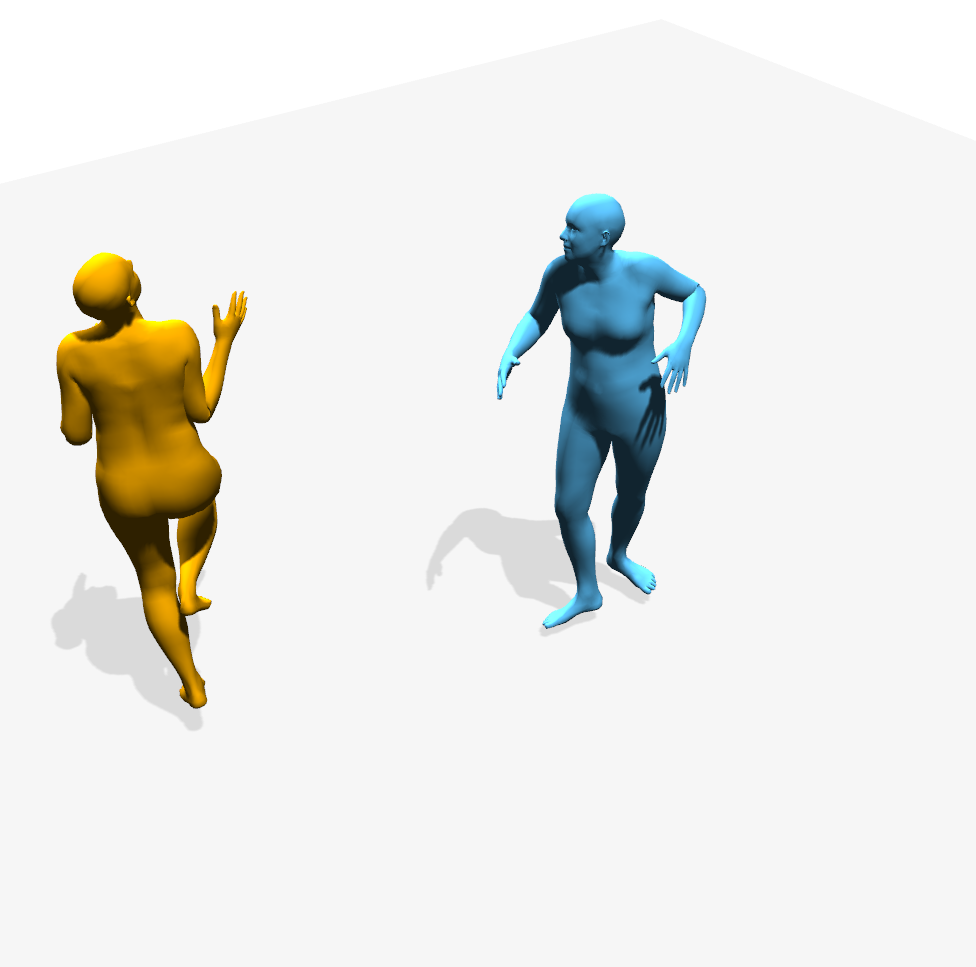} &
        \includegraphics[width=0.12\textwidth]{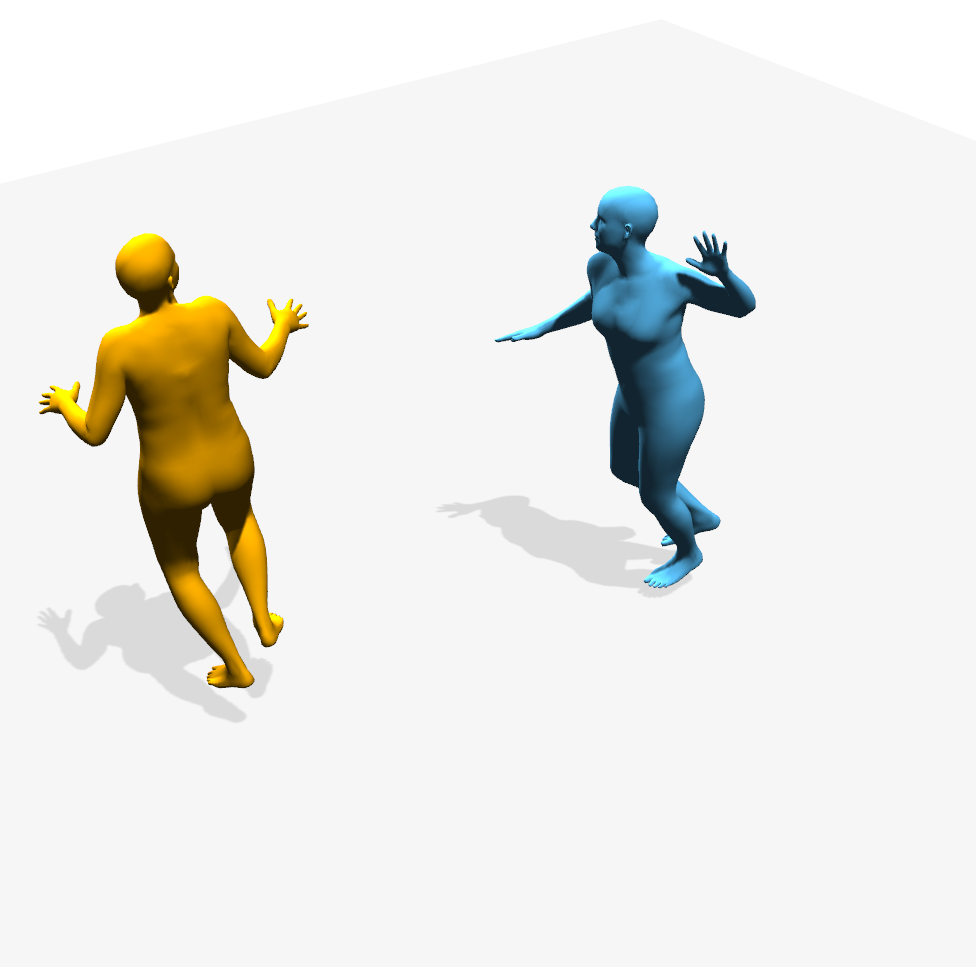} &
        \includegraphics[width=0.12\textwidth]{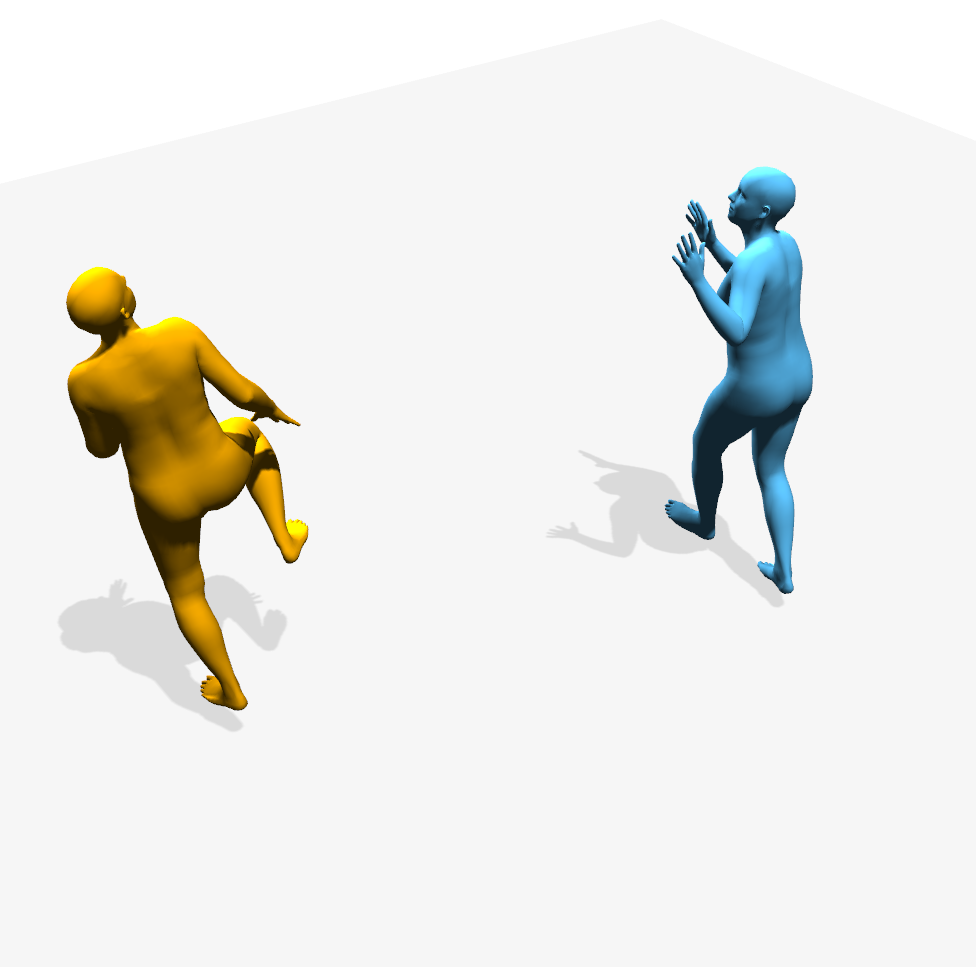} &
        \includegraphics[width=0.12\textwidth]{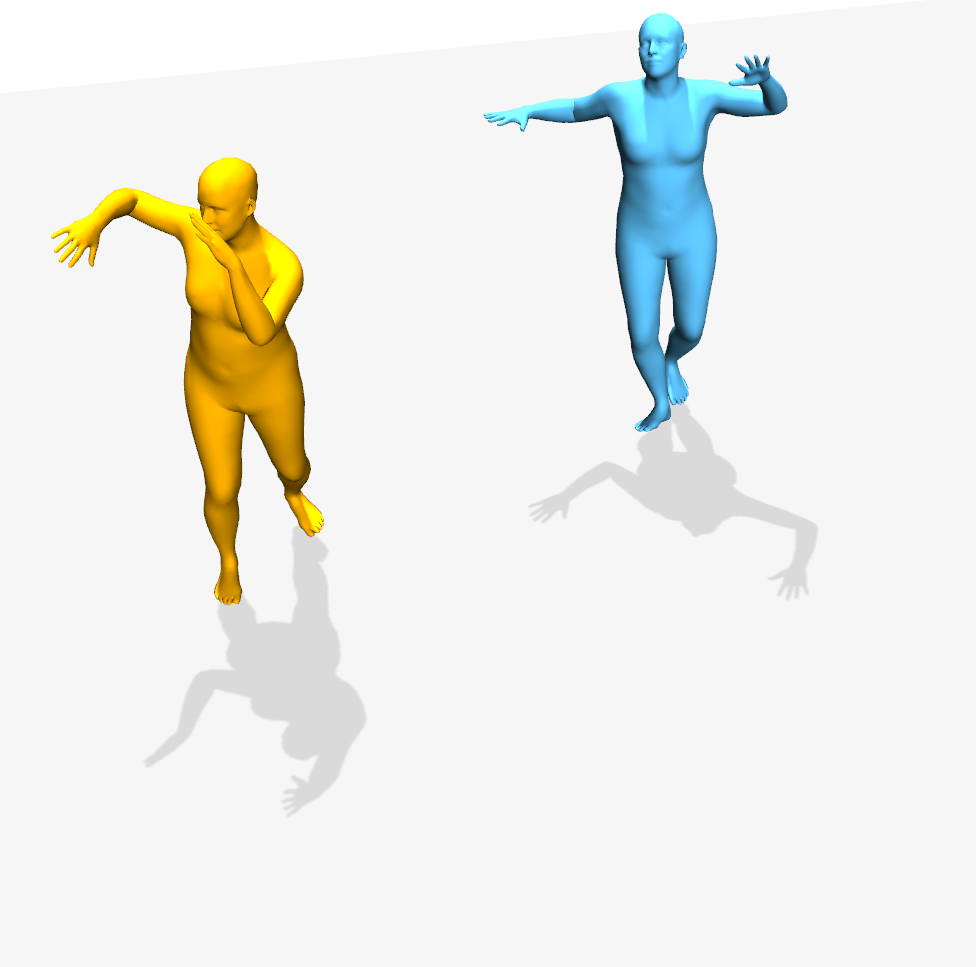} &
        \includegraphics[width=0.12\textwidth]{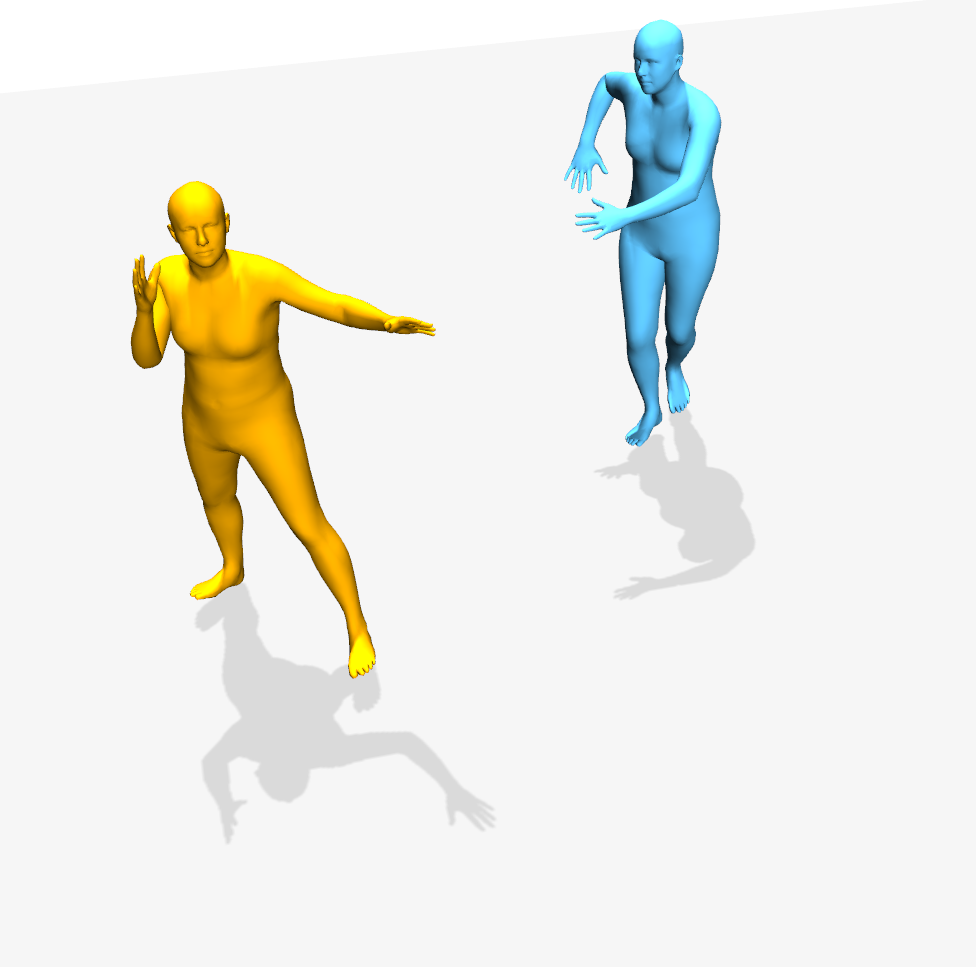} &
        \includegraphics[width=0.12\textwidth]{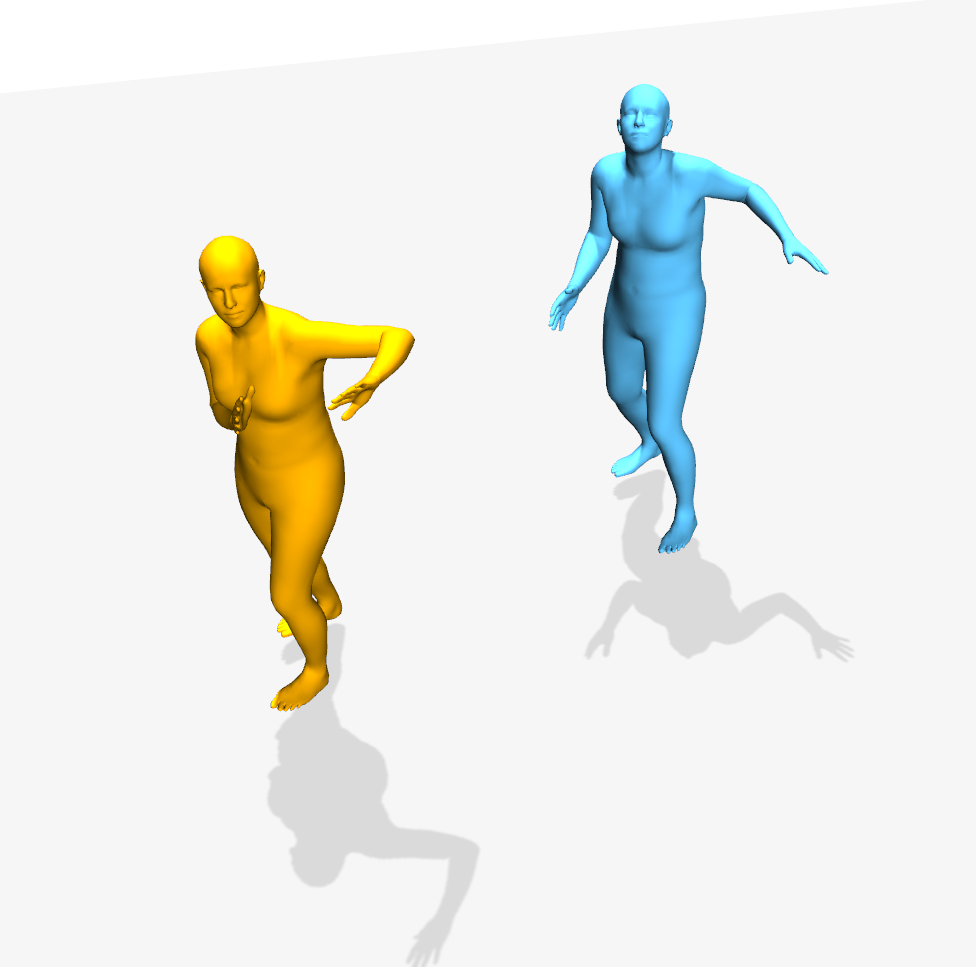} &
        \includegraphics[width=0.12\textwidth]{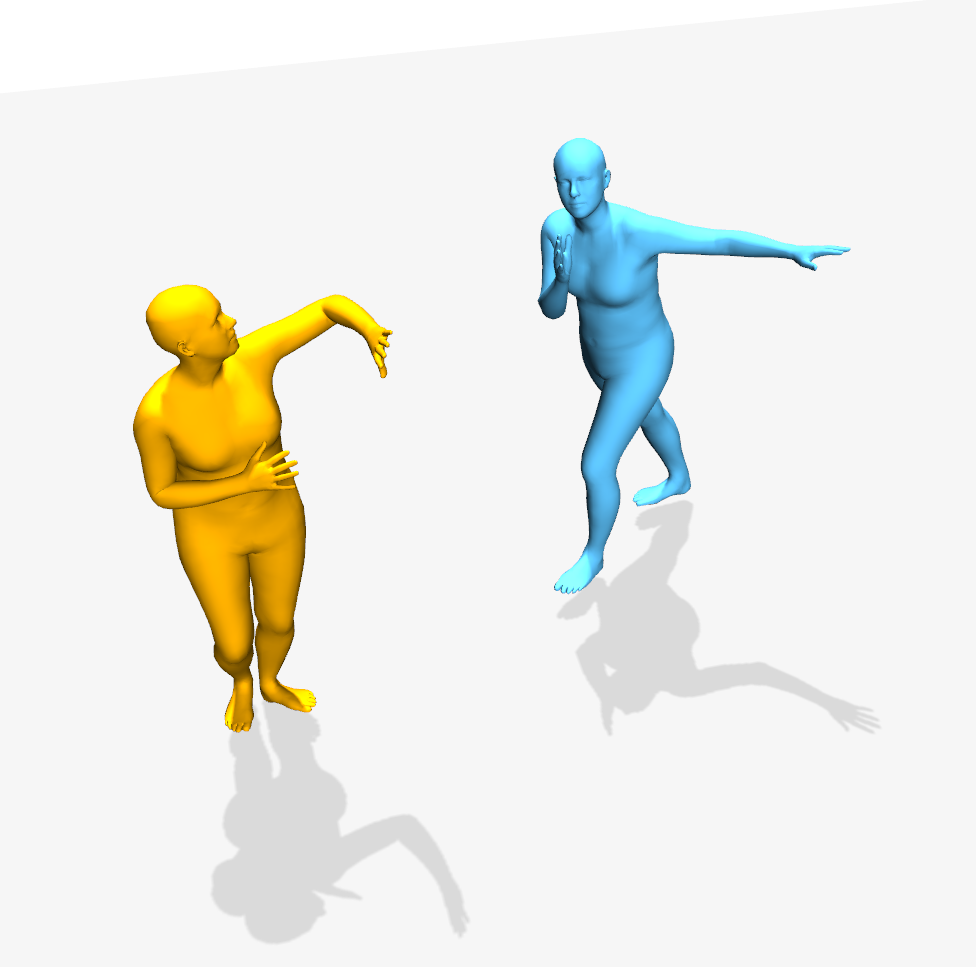}  \\ 
    \end{tabular}}

        {\itshape {The first one bumps into the second one while walking.}}
    
    {\setlength{\tabcolsep}{0pt}
    \begin{tabular}{cccc @{\hspace{2.5pt}} | @{\hspace{2.5pt}} cccc}
    
        \includegraphics[width=0.12\textwidth]{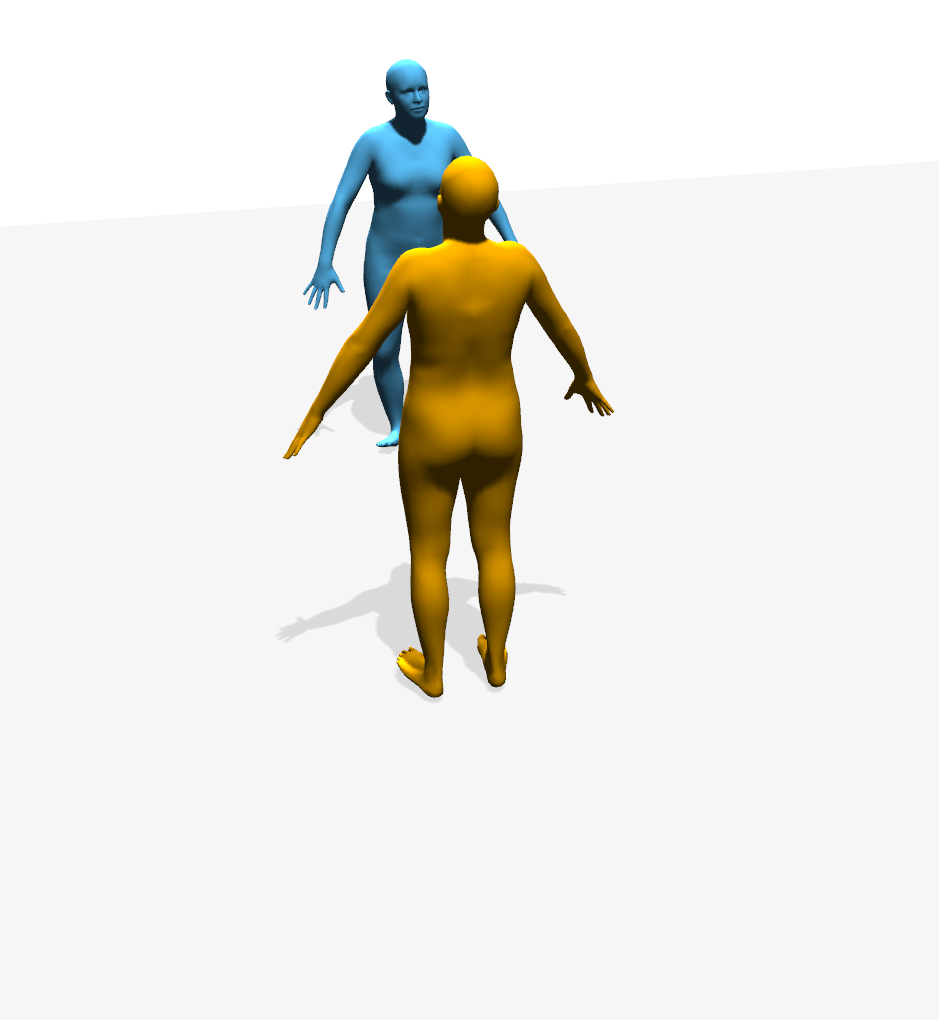} &
        \includegraphics[width=0.12\textwidth]{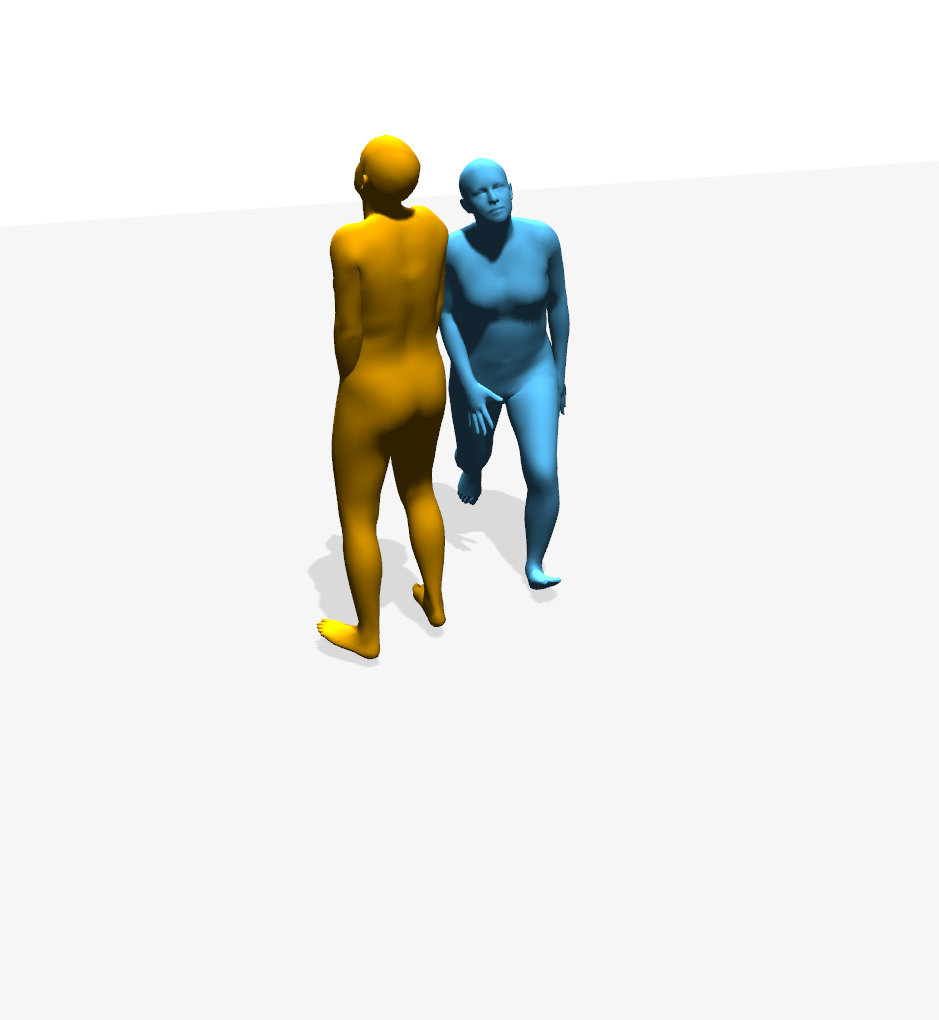} &
        \includegraphics[width=0.12\textwidth]{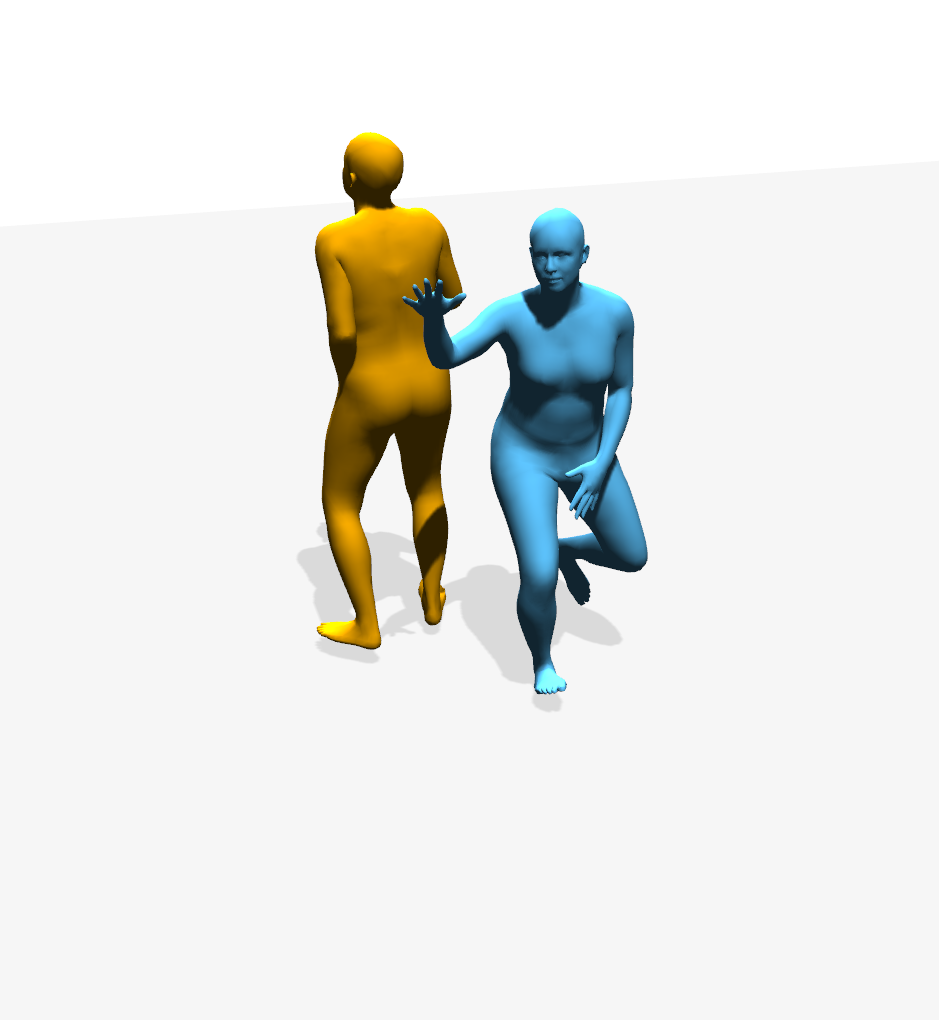} &
        \includegraphics[width=0.12\textwidth]{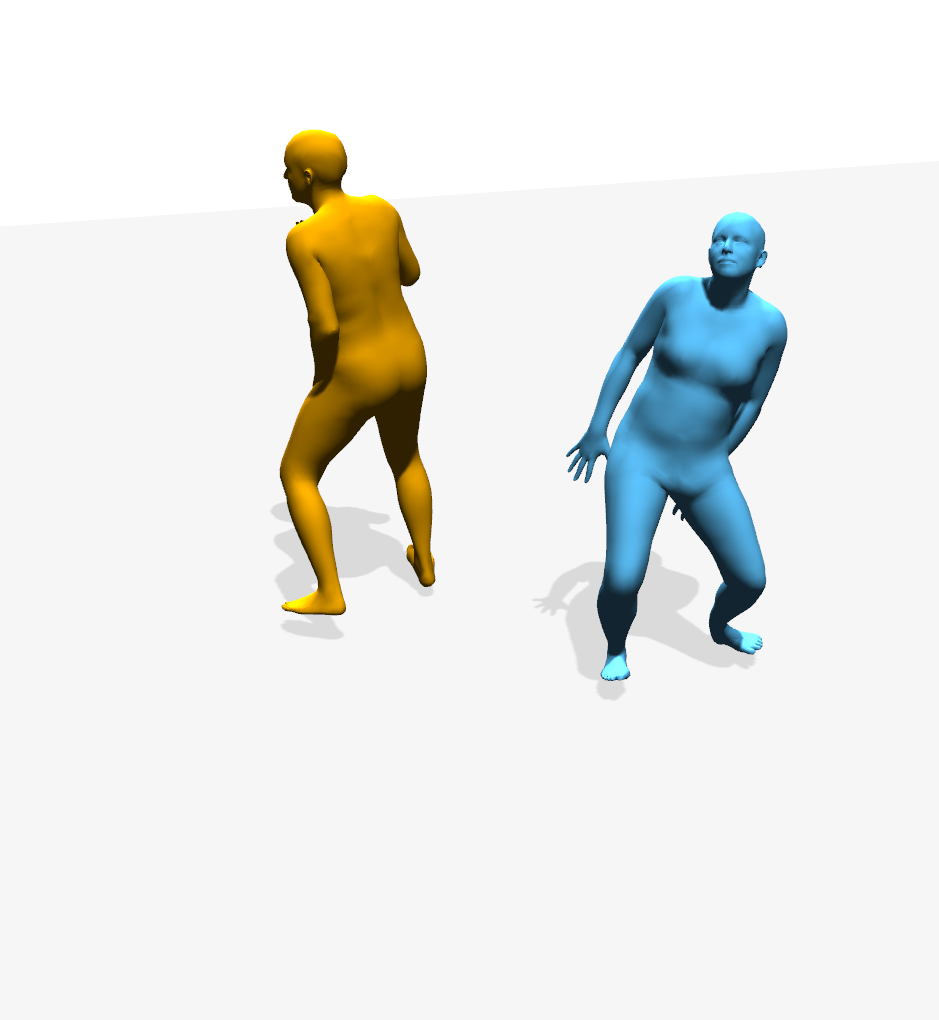} &
        \includegraphics[width=0.12\textwidth]{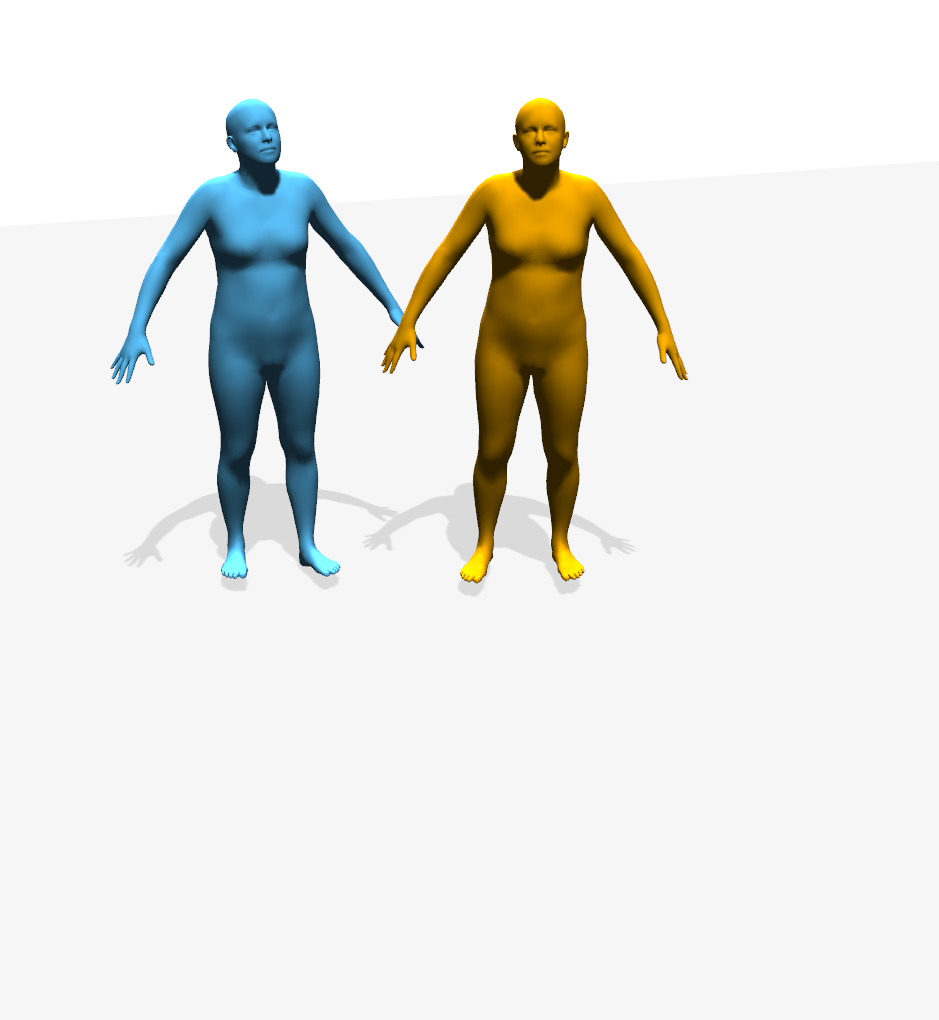} &
        \includegraphics[width=0.12\textwidth]{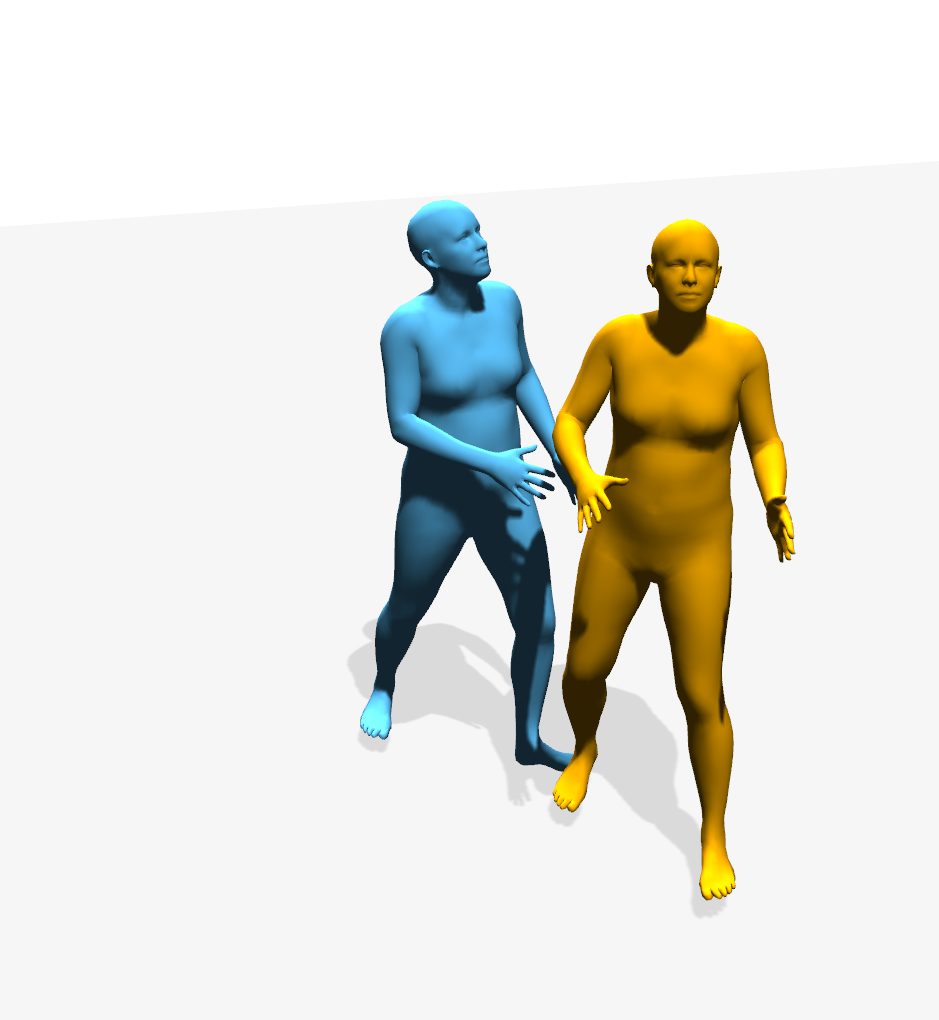} &
        \includegraphics[width=0.12\textwidth]{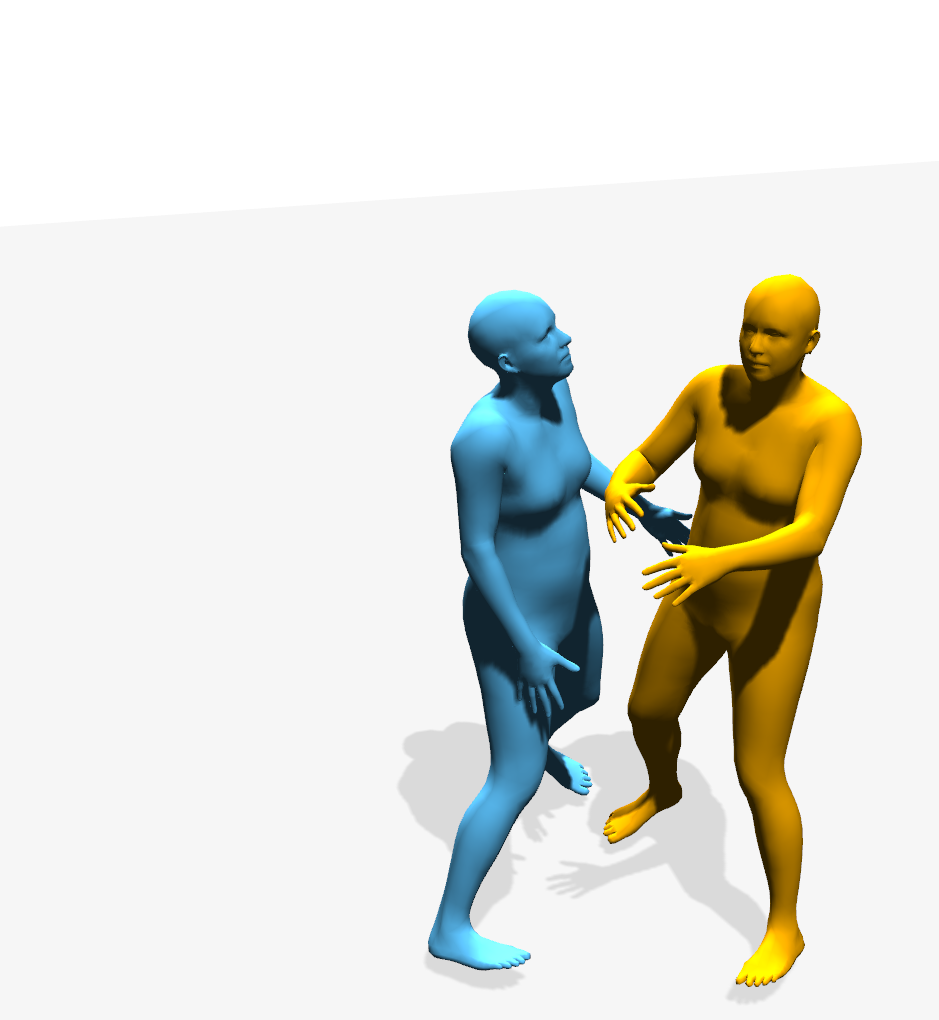} &
        \includegraphics[width=0.12\textwidth]{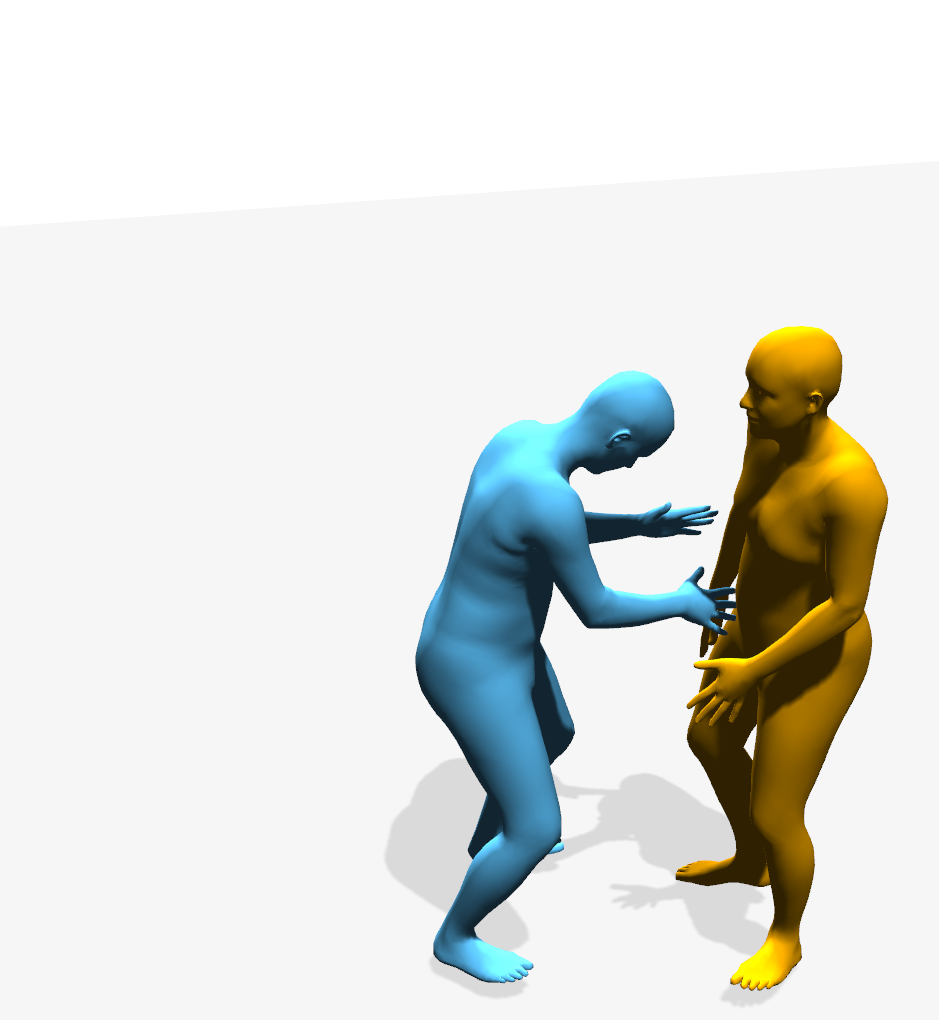}  \\ 
    \end{tabular}}
    \vspace{0.2cm}
    \caption{
        Qualitative comparison~\cite{Kaufmann_Vechev_aitviewer_2022} (zoom in to see it better) between baseline MLD++~\cite{dai2024motionlcm} and DiMoT, highlighting DiMoT's superior generation quality. (See the video in supplementary material.) 
    }

\label{qual_fig}

\end{figure*}

\subsection{Main Results}

Table~\ref{main_tab} shows DiMoT achieves state-of-the-art performance on the InterHuman~\cite{liang2024intergen} benchmark. Compared to the MLD++~\cite{dai2024motionlcm} baseline, DiMoT improves Top-3 R-Precision from 0.688 to 0.739 and reduces FID by 6.5\% (5.046 to 4.716). 
Furthermore, DiMoT consistently outperforms recent methods (e.g., DHVAE~\cite{geng2026disentangled}, TIMotion~\cite{wang2025timotion}) across all R-Precision tiers while establishing the lowest overall FID.
Similar to DHVAE, these methods based on LDMs suffer from low MModality, while our DiMoT still shows better diversity over the MLD++, with the Diversity metric closely matching the ground truth.
In the Inter-X evaluations, compared to the MLD++ baseline, DiMoT improves R-Precision (0.732 to 0.746) and achieves a substantial 24.9\% reduction in FID (0.446 to 0.335).
DiMoT still outperforms the DHVAE across R-Precision metrics, achieving the lowest FID and MM Dist.
These results confirm DiMoT's generalizability to both SMPL~\cite{SMPL:2015} and SMPL-X~\cite{SMPL-X:2019} data.
While suppressing high-frequency shortcuts with CSA slightly degrades reconstruction FID, the MPJPE improvement demonstrates enhanced kinematic accuracy.
%

%







Figure~\ref{qual_fig} qualitatively compares DiMoT with the MLD++~\cite{motionlcm-v2} baseline across diverse interactive text prompts. 
%
%
In the first scenario, MLD++ fails to capture the interaction semantics with interpenetration.
Conversely, DiMoT synthesizes the first agent's sprinting approach and the second's synchronized dismissing gesture.
For the dancing prompt, MLD++ yields generic, rigid movements that ignore the target style.
DiMoT instead generates expressive, coordinated poses, capturing complex kinematics like "stomping feet" and "swaying hips". 
In the "bumping" scenario, MLD++ poorly models the interaction, yielding unnatural spatial proximity.
DiMoT models spatial boundaries and momentum strictly with natural post-contact reactions. 
These visualization results demonstrate that DiMoT mitigates temporal artifacts, showing smooth and semantically accurate multi-person motions.
Furthermore, Appendix~\ref{app:intermask-user-study} reports a user study comparing DiMoT with InterMask.

\begin{table}[t]
\vspace{-3mm}
\caption{Ablation study of DiMoT components. Results are from the InterHuman~\cite{liang2024intergen} dataset.
}
\centering
\resizebox{1\columnwidth}{!}{
\begin{tabular}{lcccccc}
\toprule
\multirow{2}{*}{Method} & \multicolumn{3}{c}{Tokenizer} & \multicolumn{3}{c}{Diffusion} \\
\cmidrule(lr){2-4} \cmidrule(lr){5-7}
& rFID $\downarrow$ & rR-Top3 $\uparrow$ & MPJPE $\downarrow$ & FID $\downarrow$ & R-Top3 $\uparrow$ & Foot Sliding $\downarrow$ \\
\midrule
MLD~\cite{chen2023executing}    &   0.317   & 0.629   & 71.10 & 5.477    &  0.636   &    0.0209   \\
GMM~\cite{tuautoregressive}    &   0.183   & 0.651   & 58.15 & 5.164    &  0.665   &    0.0191   \\
MLD++~\cite{motionlcm-v2} &   0.080   & 0.686   & 49.26 & 5.046    &  0.688   &    0.0186   \\
\bottomrule
MLD~\cite{chen2023executing} w. CSA  & 0.315  & 0.661  & 67.70  & 5.322  & 0.689    & 0.0203 \\
MLD~\cite{chen2023executing} w. AMD  & 0.222  & 0.656  & 59.34  & 5.209  & 0.662    & 0.0194 \\
\rowcolor{orange!15} MLD~\cite{chen2023executing} w. DiMoT & \textbf{0.193} {\scriptsize \textcolor{green!60!black}{(-0.124)}} & \textbf{0.663} {\scriptsize \textcolor{green!60!black}{(+0.034)}} & \textbf{53.90} {\scriptsize \textcolor{green!60!black}{(-17.20)}} & \textbf{5.245} {\scriptsize \textcolor{green!60!black}{(-0.232)}} & \textbf{0.685} {\scriptsize \textcolor{green!60!black}{(+0.049)}} & \textbf{0.0197} {\scriptsize \textcolor{green!60!black}{(-0.0012)}} \\
\bottomrule
MLD++~\cite{motionlcm-v2} w. CSA    & 0.086    & 0.713  & 47.18  & 4.967  & 0.746   & 0.0182 \\
MLD++~\cite{motionlcm-v2} w. AMD    & 0.077    & 0.672  & 46.76  & 4.781  & 0.716   & 0.0175 \\
\rowcolor{orange!15} MLD++~\cite{motionlcm-v2} w. DiMoT & \textbf{0.071} {\scriptsize \textcolor{green!60!black}{(-0.009)}} & \textbf{0.698} {\scriptsize \textcolor{green!60!black}{(+0.012)}} & \textbf{41.36} {\scriptsize \textcolor{green!60!black}{(-7.90)}} & \textbf{4.716} {\scriptsize \textcolor{green!60!black}{(-0.330)}} & \textbf{0.739} {\scriptsize \textcolor{green!60!black}{(+0.051)}} & \textbf{0.0168} {\scriptsize \textcolor{green!60!black}{(-0.0018)}} \\
\bottomrule
\label{tab_dimot_abla}
\vspace{-0.5cm}
\end{tabular}}
\end{table}


\begin{figure*}[t]
    \centering

    \begin{subfigure}[t]{0.48\textwidth}
        \centering
        \includegraphics[width=\linewidth]{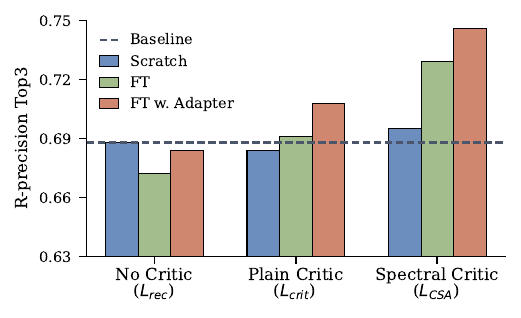}
        \caption{R-precision results for different encoder designs. FT refers to fine-tuning without extra parameters.}
        \label{fig:csa_bar_a}
    \end{subfigure}
    \hfill
    \begin{subfigure}[t]{0.48\textwidth}
        \centering
        \includegraphics[width=\linewidth]{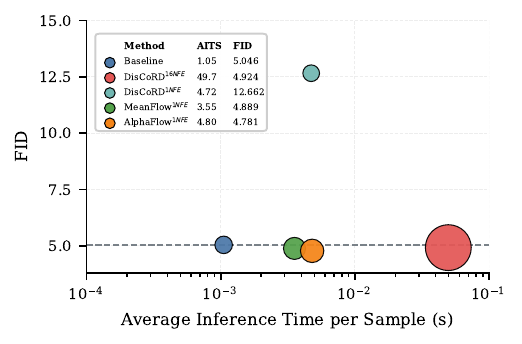}
        \caption{FID results for different decoder designs.
        We use the circle size to represent the decoder FLOPs.
        }
        \label{fig:csa_bar_b}
    \end{subfigure}

    \caption{Comparison of different design choices with DiMoT on the InterHuman dataset.}
    \label{fig:csa_bar_pair}
\vspace{-3mm}
\end{figure*}

\subsection{Ablation Study}
Table~\ref{tab_dimot_abla} ablates the DiMoT components across Latent Diffusion Model (LDM) baselines (MLD~\cite{chen2023executing}, MLD++~\cite{motionlcm-v2}). 
Integrating DiMoT consistently improves generation quality independent of the base architecture. 
By applying our tokenizer design to MLD++, DiMoT achieves the best generative performance with minimum temporal errors (Foot Sliding: 0.0168).
DiMoT improves the diffusability with a superior trade-off between reconstruction and generation.
The tokenizer maintains its fidelity with stronger MPJPE results while maximizing motion fidelity and text alignment.

%
Figure~\ref{fig:csa_bar_pair}(a) compares different regularization methods applied to the tokenizer encoder.
"No Critic" refers to the encoder that relies solely on a reconstruction loss, which evaluates the effectiveness of the reconstruction objective itself.
%
%
CSA achieves optimal performance (R-Top3: 0.74) when fine-tuned with an additional adapter, which is further enhanced by spectral reweighting.
Without an adapter, CSA functions as a parameter-free objective applied to the standard MLD++~\cite{motionlcm-v2} diffusion backbone, yet achieves better text-motion alignment. More spectral reweighting results are included in Appendix~\ref{csa_results}.

Figure~\ref{fig:csa_bar_pair}(b) evaluates various decoder generative schedulers in terms of latency and FID.
We also highlight the inference step and FLOPs with the superscript and circle size, respectively.
DisCoRD$^{16\text{NFE}}$ achieves better fidelity than the baseline but with excessive latency and FLOPs.
In contrast, our AMD improves the generation fidelity with single-step efficiency, reducing computational overhead by orders of magnitude (See more results in Appendix~\ref{AMD_eval}).
Moreover, we compare linear and sigmoid curricula in Appendix~\ref{app:alpha-schedule}.

\begin{wraptable}{r}{0.5\columnwidth}
\vspace{-4mm} 
\caption{Single-person results on HumanML3D.}
\label{tab:single_extra}
\centering
\resizebox{1\linewidth}{!}{
\begin{tabular}{l|cccc}
\toprule
Method & R-Top3$\uparrow$ & FID$\downarrow$ & MM Dist$\downarrow$ & MModality$\uparrow$ \\
\midrule
GT          & $0.511$ & $0.002$ & $2.974$ & - \\
\bottomrule
MLD++~\cite{motionlcm-v2}       & $0.544$ & $0.049$ & $2.828$ & $1.672$ \\
MLD++~\cite{motionlcm-v2} w. CSA & $0.552$ & $0.047$ & $2.808$ & $1.715$ \\
MLD++~\cite{motionlcm-v2} w. AMD & $0.542$ & $0.042$ & $2.842$ & $1.675$ \\
MLD++~\cite{motionlcm-v2} w. DiMoT & $0.550$ & $0.038$ & $2.809$ & $1.689$ \\
\bottomrule
\end{tabular}}
\vspace{-2mm}
\end{wraptable}

In Table~\ref{tab:single_extra}, we also conduct an evaluation on the single-person benchmark (\ie, HumanML3D~\cite{guo2022generating}).
By integrating DiMoT into MLD++~\cite{motionlcm-v2}, it also improves the baseline performance.
DiMoT dominates in fidelity with the lowest FID, and also shows gains in R-Top3 and MModality.
These results further confirm that DiMoT's architectural benefits generalize to motion domains.

\vspace{-0.3cm}
\section{Conclusion}

In this paper, we address the generation-reconstruction trade-off in motion latent diffusion models by improving diffusability of the tokenizer, rather than scaling the diffusion backbone.
We observe a spectral mismatch wherein both the latent and output profiles exhibit an abnormally high proportion of high-frequency components, potentially degrading text-to-motion generation.
To address this, we propose the Diffusable Motion Tokenizer (DiMoT), consisting of Critic Spectral Adaptation (CSA) and Alpha-Flow Motion Decoder (AMD).
Firstly, we demonstrate, both theoretically and empirically, that CSA regularizes the latent space via a diffusion-based spectral critic, suppressing high-frequency shortcuts without sacrificing reconstruction.
Concurrently, AMD replaces deterministic decoding with a one-step average-velocity prediction, selectively preserving the essential dynamics required for temporal realism.
Evaluations on standard motion benchmarks demonstrate that DiMoT successfully balances generation fidelity and reconstruction accuracy, validating our tokenizer refinement strategy.
%


\textbf{Limitation}
DiMoT addresses spectral mismatch in text-to-motion generation, but defining the optimal spectral profile of the latent space remains an open problem, where CSA relies on using the diffusion as a critic with empirical choices.
Moreover, motion datasets are limited in scale and use heterogeneous representations.
Our evaluations demonstrate improvements within each dataset, but do not establish whether a tokenizer adapted on one dataset retains its performance on another.
%
Future work will explore reinforcement or adversarial learning to replace the polynomial spectral weight, bypassing these rigid priors and optimality assumptions.
%


\section*{Acknowledgments}
This work was supported by the K-CSC funding.
The authors acknowledge the use of King’s CREATE HPC. Retrieved November 05, 2026, from \url{https://doi.org/10.18742/rnvf-m076}.

\newpage







{
    \small
    \bibliographystyle{unsrtnat}
    \bibliography{main}
}


\appendix



\section{Full theoretical development of CSA}\label{app:csa-full}

\subsection{Notation}\label{app:notation}

\begin{center}
\small
\begin{tabular}{ll}
\toprule
Symbol & Meaning \\
\midrule
$\X,\Latents$            & data and latent spaces, $\Latents=\R^d$ with $d=C'\times N$, $N=n_1\cdots n_D$ \\
$E_\ph,D_{\psi_0}$       & encoder (with parameters $\ph$) and decoder pinned at its pre-trained parameters $\psi_0$ \\
$\pdata$                 & data distribution \\
$\pz^\ph$                & induced latent distribution $(E_\ph)_\#\pdata$ \\
$f_\theta,\,\theta_0$    & velocity model (rectified flow) and its frozen pre-trained parameters \\
$\sg(\cdot)$             & stop-gradient (identity on values, zero on gradients) \\
$x_t,\epsilon,t$         & rectified-flow interpolant, noise, time \\
$y,\,p_y,\,p_{z|y}^\ph$  & text prompt, text marginal, conditional latent law $(E_\ph)_\#p(x|y)$ \\
$r(z)$                   & prediction residual $\sg(\vpred)-(\epsilon-z)$, with $\vpred\!\coloneqq\!f_{\theta_0}(x_t,t,y)$ \\
$\Lcrit$                 & plain critic loss \\
$\Lcsa$                  & static spectrally weighted CSA loss \\
$\Lrec$                  & reconstruction loss \\
$\Ldiff(\theta;\ph)$     & diffusion training loss for encoder $\ph$ \\
$\FFT,\FFT^*$ & unitary $D$-D Discrete Fourier Transform (DFT) and its conjugate transpose \\
$d(k)$                   & normalized Euclidean FFT distance from DC, $d(k)\in[0,1]$ \\
$w(k),\tilde w(k),\bar w$& static spectral weight, mean-1-normalized weight, mean of $w$ \\
$\kappa,p$               & spectral hyperparameters: HF strength and distance exponent \\
$\lambda_c$              & weight on the spectral critic loss \\
$M_w$                    & spectral mass operator $\FFT^{*}\diag(\tilde w)\FFT$ \\
$N$                      & total spatial DFT modes $n_1\cdots n_D$ ($N=T$ for the 1D motion setting of Section~\ref{sec:spec-critic}) \\
$H$                      & $\Lrec$ Hessian at the baseline $\ph_0$ \\
$g$                      & stop-gradient pseudo-gradient of $\Lcsa$ at $\ph_0$ (\Cref{asm:critic-smooth}) \\
$g_{\mathrm{crit}}$      & true gradient $\nabla_\ph\Ldiff(\theta_0;\ph_0)$ at the baseline (not the stop-gradient pseudo-gradient) \\
$\PSD(k)$                & per-mode power spectral density of the latent process \\
$\beta_{\mathrm{eff}}(k,t)$ & per-mode inverse effective temperature (\Cref{prop:freq-temp}) \\
\bottomrule
\end{tabular}
\end{center}

\subsection{Setup}\label{app:setup}

\begin{definition}[Stop-gradient operator]\label{def:sg}
The stop-gradient $\sg:\R^N\to\R^N$ is the identity map on values: $\sg(z)=z$ as elements of $\R^N$. In any expression where $\sg(z(\ph))$ appears, gradients with respect to $\ph$ treat $\sg(z(\ph))$ as a constant. Equivalently, introduce a frozen copy $\bar z=z$ with $\partial\bar z/\partial\ph\equiv 0$. Replace $\sg(z(\ph))$ by $\bar z$ in every gradient computation. Operationally, in PyTorch this is \texttt{z.detach()}.
\end{definition}

Let $\X=\R^{C\times H\times W}$ and $\Latents=\R^d$ be the data and latent spaces. Set $d=C'\times n_1\times\cdots\times n_D$ with $D\in\{1,2,3\}$ for motion, image, and video respectively. An autoencoder is a pair $(E_\ph,D_{\psi_0})$ with text-agnostic encoder $E_\ph(x)$ and a frozen pre-trained decoder $D_{\psi_0}$. Data is sampled jointly $(x,y)\sim\pdata$ where $y$ is the text prompt, with text marginal $p_y$. The conditional latent law is $p_{z|y}^\ph\coloneqq(E_\ph)_\#p(x|y)$, and the marginal latent law is $\pz^\ph=(E_\ph)_\#\pdata$. The rectified-flow interpolant~\citep{liu2022flow} is $x_t=(1-t)z_0+t\epsilon$ with velocity $v=\epsilon-z_0$. A text-conditioned velocity model $f_\theta(x_t,t,y)$ is trained by
\begin{equation}\label{eq:diff-loss-app}
  \Ldiff(\theta;\ph)
  = \E_{(x,y),\,t\sim\mathcal{U}[0,1],\,\epsilon\sim\N(0,I)}
        \norm{f_\theta((1-t)E_\ph(x)+t\epsilon,\,t,\,y)-(\epsilon-E_\ph(x))}^2.
\end{equation}

\begin{definition}[Plain critic loss]\label{def:plain-critic}
For $z=E_\ph(x)$, sample $(t,\epsilon)$ as above, set $x_t=(1-t)\sg(z)+t\epsilon$, and define the prediction residual $r(z)=\sg(f_\theta(x_t,t,y))-(\epsilon-z)$. The plain critic loss is
\begin{equation}\label{eq:plain-crit-app}
  \Lcrit(\ph;\theta) = \E_{(x,y),t,\epsilon}\norm{r(z)}^2.
\end{equation}
\end{definition}

\begin{lemma}[Value equivalence (values, not gradients)]\label{lem:value-equiv}
For every $\ph$ and every frozen $\theta$, $\Lcrit(\ph;\theta)=\Ldiff(\theta;\ph)$ as values. Their gradients in $\ph$ differ under the stop-gradients (\Cref{rem:value-vs-grad}, \Cref{asm:critic-smooth}).
\end{lemma}

\begin{proof}
$\sg$ preserves values, so dropping $\sg$ from \eqref{eq:plain-crit-app} gives
\[
  \E_{(x,y),t,\epsilon}\norm{f_\theta\!\bigl((1-t)E_\ph(x)+t\epsilon,\,t,\,y\bigr)-\epsilon+E_\ph(x)}^2,
\]
which matches \eqref{eq:diff-loss-app} with $z_0=E_\ph(x)$. Equivalently, by the tower property $\E_{(x,y)}=\E_y\E_{x|y}$, the same quantity reads $\E_{y\sim p_y}\E_{z_0\sim p_{z|y}^\ph,\,t,\,\epsilon}\norm{f_\theta(x_t,t,y)-(\epsilon-z_0)}^2$, the form used in conditional diffusion training.
\end{proof}

\begin{remark}[Stop-gradients: value equivalence and necessity of both stops]\label{rem:value-vs-grad}
$\Lcrit$ and $\Ldiff$ agree as values (\Cref{lem:value-equiv}). Their gradients in $\ph$ differ under the stop-gradients. Autograd of $\Lcrit$ drops the Jacobian-of-$f_{\theta_0}$ contribution that $\nabla_\ph\Ldiff$ retains. The pseudo-gradient $g$ (\Cref{asm:critic-smooth}) is the autograd direction. The true $g_{\mathrm{crit}}=\nabla_\ph\Ldiff(\theta_0;\ph_0)$ is the Taylor coefficient in \Cref{thm:diffusion-bound}.

Both stops are required. Removing $\sg$ from $\vpred$ adds a $J_{f_{\theta_0}}^\top J_{f_{\theta_0}}$ term to the Hessian and breaks \Cref{thm:strong-convex}. Removing $\sg$ from $x_t$ instead pushes the encoder to reshape the noised sample itself, a different objective.
\end{remark}

\subsection{FFT distance grid and CSA loss}\label{app:dist}

\begin{definition}[Normalized FFT distance grid]\label{def:dist}
For spatial sizes $(n_1,\dots,n_D)$ let $f^{(\ell)}_{k_\ell}=k_\ell/n_\ell\pmod 1\in[-1/2,1/2)$ in FFT layout. Define
\[
  d(k_1,\dots,k_D) = \Bigl(\textstyle\sum_\ell (f^{(\ell)}_{k_\ell})^2\Bigr)^{1/2}\!/d_{\max},
  \qquad
  d_{\max} = \max_k\Bigl(\textstyle\sum_\ell (f^{(\ell)}_{k_\ell})^2\Bigr)^{1/2}.
\]
Then $d(0)=0$ at DC, and $\max_k d(k)=1$ by normalization. The maximum is attained at the Nyquist corner $k_\ell=n_\ell/2$ when all $n_\ell$ are even. Moreover, $d(\bar k)=d(k)$ where $\bar k_\ell\coloneqq n_\ell-k_\ell$ is the per-axis reflection, since $f^{(\ell)}_{n_\ell-k_\ell}\equiv -f^{(\ell)}_{k_\ell}\pmod 1$ in the FFT layout.
\end{definition}

\begin{definition}[CSA loss]\label{def:spec-critic}
With $w(k)=1+\kappa\,d(k)^p$ and $\tilde w(k)=w(k)/\bar w$, $\bar w=N^{-1}\sum_k w(k)$, and $\FFT$ the unitary $D$-D DFT acting on the spatial axes,
\[
  \Lcsa(\ph;\theta,w)
  = \E_{(x,y),t,\epsilon}\!\Bigl[\tfrac{1}{N}\sum_k \tilde w(k)\bigl\lvert(\FFT r)(k)\bigr\rvert^2\Bigr].
\]
Here $\bigl\lvert(\FFT r)(k)\bigr\rvert^2 = \sum_c \bigl\lvert(\FFT r)(k,c)\bigr\rvert^2$ sums the squared magnitude over channels at each spatial mode $k$, and $N=\prod_\ell n_\ell$ is the number of spatial modes.
\end{definition}

\begin{lemma}[Parseval recovery]\label{lem:parseval}
With $\kappa=0$, so $w\equiv\tilde w\equiv 1$, $\Lcsa(\ph;\theta,1)=\Lcrit(\ph;\theta)/N$.
\end{lemma}

\begin{proof}
Plancherel for the unitary DFT gives $\sum_k|(\FFT u)(k)|^2=\sum_i u_i^2$ for any $u\in\R^N$. Apply per channel to $u=r$.
\end{proof}

\subsection{Pointwise gradient, Hessian, strong convexity}\label{app:struct}

\begin{proposition}[Pointwise gradient]\label{prop:grad}
Fix $(x,y,t,\epsilon)$ and let $c=\sg(f_\theta(x_t,t,y))$, constant in $z$ under stop-gradient. Let $M_w=\FFT^{*}\diag(\tilde w)\FFT\in\R^{N\times N}$. Then $M_w^\top=M_w$, $M_w$ maps $\R^N$ to $\R^N$, and
\begin{enumerate}[label=(\roman*),leftmargin=*,nosep]
\item $\ell_w(z)\coloneqq\tfrac{1}{N}\sum_k \tilde w(k)|(\FFT r(z))(k)|^2=\tfrac{1}{N}\,r(z)^\top M_w r(z)$, with $r(z)=c-\epsilon+z$.
\item $\nabla_z\ell_w=\tfrac{2}{N}M_w r(z)=\tfrac{2}{N}M_w(z-(\epsilon-c))$.
\item $\nabla_z^2\ell_w=\tfrac{2}{N}M_w$, constant in $z$.
\item $\nabla_\ph\ell_w=\tfrac{2}{N}J_{E_\ph}(x)^\top M_w r(z)$.
\end{enumerate}
\end{proposition}

\begin{proof}
$\diag(\tilde w)$ is real-diagonal. By \Cref{def:dist}, $d(\bar k)=d(k)$, so $\tilde w(\bar k)=\tilde w(k)$ and $\tilde w$ is conjugate-symmetric in the FFT layout. For $u\in\R^N$, $\hat u\coloneqq\FFT u$ satisfies the Nyquist symmetry $\hat u(\bar k)=\overline{\hat u(k)}$. Then $\diag(\tilde w)\hat u$ inherits the symmetry and its inverse DFT is real, so $M_w u\in\R^N$. The unitary conjugation $M_w=\FFT^*\diag(\tilde w)\FFT$ of a real diagonal is Hermitian on $\C^N$. Restricting to $\R^N$ gives $M_w^\top=M_w$ with spectrum $\{\tilde w(k)\}_k$. For the multi-channel latent $r\in\R^{C'\times N}$, the same operator applies along each channel with overall Hessian $I_{C'}\otimes M_w$.

For (i), $\sum_k\tilde w(k)|(\FFT r)(k)|^2=\langle\FFT r,\diag(\tilde w)\FFT r\rangle=r^\top M_w r$ by unitarity. For (ii) and (iii), $\partial r/\partial z=I$ and $M_w$ is constant, so $\nabla_z(r^\top M_w r)/N=2M_w r/N$ and $\nabla_z^2\ell_w=2M_w/N$. For (iv), apply the chain rule at $z=E_\ph(x)$.
\end{proof}

\begin{theorem}[Strong convexity of the detached surrogate in $z$]\label{thm:strong-convex}
Let $L_w(z)=\E_{t,\epsilon}[\ell_w(z)]$ at fixed $(x,y)$, where $\ell_w$ is the pointwise loss of \Cref{prop:grad} with $c=\sg(f_\theta(x_t,t,y))$ frozen via \Cref{def:sg}. Then $\nabla_z^2 L_w=2M_w/N$, and $L_w$ is $\mu$-strongly convex and $\Lambda$-smooth in $z$ with
\[
  \mu = 2N^{-1}\min_k\tilde w(k),
  \qquad
  \Lambda = 2N^{-1}\max_k\tilde w(k).
\]
For $w(k)=1+\kappa\,d(k)^p$ with $\kappa\geq 0$, $\tilde w>0$, so $\mu>0$ and the global minimizer $z^*$ is unique. Gradient descent on $L_w$ at fixed $c$ with $\eta\leq 1/\Lambda$ contracts in the $M_w$-norm at rate $1-\eta\mu$.
\end{theorem}

\begin{remark}[Strong convexity is a frozen-$c$ statement]\label{rem:strong-conv-scope}
\Cref{thm:strong-convex} bounds the Hessian of the detached surrogate at fixed $c$. Real training re-evaluates $f_{\theta_0}$ at each step, so $c$ moves with $z$ and contributes a non-constant Jacobian term to the full Hessian. The contraction rate $1-\eta\mu$ is therefore a property of the surrogate, not a guarantee on the full training dynamics. Training also acts on the encoder parameters $\ph$, not on $z$ directly. The $z$-space convexity at fixed $(x,y,c)$ does not transfer to a corresponding property in $\ph$.
\end{remark}

\begin{proof}
\Cref{prop:grad}(iii) gives a constant Hessian at every $(x,y,t,\epsilon)$, and linearity of expectation preserves it. $M_w$ is the unitary conjugation of a positive diagonal, so its spectrum is exactly $\{\tilde w(k)\}_k$, giving $\mu$ and $\Lambda$. Strict positivity of $\tilde w$ gives a unique minimizer. The contraction statement is the standard rate for quadratic objectives with constant Hessian in the $M_w$-norm.
\end{proof}

\begin{remark}[Fixed-point set is invariant under $\tilde w$]\label{rem:fixed-point}
$z^*$ minimizes $L_w$ at fixed detached $c=\sg(\vpred)$ if and only if $\E_{t,\epsilon}[c-(\epsilon-z^*)]=0$. This follows from setting $\nabla_z L_w=0$ in \Cref{prop:grad}(ii) and inverting $M_w$. The fixed-point set is therefore the same under any positive weighting. Spectral weights steer the dynamics toward this set but do not change which $z^*$ are stationary. Plain critic and CSA share the fixed points. What differs is the gradient field at non-stationary $z$. The practical benefit of CSA therefore arises on the way to convergence, not from changing the asymptotic optimum.
\end{remark}

\begin{theorem}[Per-frequency selectivity]\label{thm:spectral-grad}
Let $\hat r=\FFT r$. The CSA gradient in the Fourier domain is
\[
  (\FFT\,\nabla_z\ell_w)(k) = \tfrac{2}{N}\tilde w(k)\,\hat r(k),
\]
and the per-mode magnitude is $2\tilde w(k)|\hat r(k)|/N$. The gradient at mode $k$ vanishes if and only if $\tilde w(k)=0$ or $\hat r(k)=0$.
\end{theorem}

\begin{proof}
Apply $\FFT$ to \Cref{prop:grad}(ii) and use unitarity $\FFT\FFT^*=I$, so $\FFT M_w=\diag(\tilde w)\FFT$.
\end{proof}

\subsection{Bayes-optimal sanity check}\label{app:self-consist}

This subsection is stated for the marginal latent law $\pz^{\ph_0}$ to keep the analysis self-contained. The conditional case follows by the same argument with $p_y\E_{x|y}$ replacing $\E_x$ and $\Delta(z,y)$ replacing $\Delta(z)$, but is not used by \Cref{thm:diff-loss}.

\begin{remark}[Spring-force interpretation]\label{rem:spring}
The pointwise gradient $\nabla_z\ell_w=2N^{-1}M_w(z-(\epsilon-\vpred))$ acts as a spring pulling $z$ toward $\epsilon-\vpred$, the diffusion's implicit reconstruction of the clean code from $x_t$. Stiffness $M_w$ upweights the modes selected by $\tilde w$. The fixed point in \Cref{rem:fixed-point} is the configuration where encoder outputs and diffusion denoising agree on every mode.
\end{remark}

\begin{proposition}[Bayes-optimal sanity check]\label{prop:bayes-opt}
Let $f^*(x_t,t)\coloneqq\E[\epsilon-z_0\mid x_t,t]$ be the Bayes-optimal velocity predictor for $\pz^{\ph_0}$. Define the population residual
\[
  \Delta(z) \coloneqq \E_{t,\epsilon}[f^*((1-t)z+t\epsilon,t)-(\epsilon-z)],
\]
and the per-time slice
\[
  \Delta_t(z) \coloneqq \E_\epsilon[f^*((1-t)z+t\epsilon,t)-(\epsilon-z)],
\]
so $\Delta(z)=\E_t[\Delta_t(z)]$. Then $\E_{z\sim\pz^{\ph_0}}[\Delta(z)]=0$. Pointwise, $\Delta(z)\neq 0$ in general. For $z\in\mathrm{supp}(p_z)$ with $p_z\in C^2$ and $p_z(z)>0$, $\Delta_t(z)\to 0$ as $t\to 0^+$ at rate $O(t)$, uniformly on compact subsets of $\mathrm{supp}(p_z)$.
\end{proposition}

The vanishing of $\E_z[\Delta(z)]$ is a sanity check on the residual: its mean over the training distribution is zero. It does not assert encoder-parameter stationarity. The actual encoder gradient at $\ph_0$ is, by \Cref{prop:grad}(iv),
\[
  \nabla_\ph\Lcsa(\ph_0;\theta_0,w) = \tfrac{2}{N}\,\E_x\bigl[J_{E_{\ph_0}}(x)^\top M_w\,\Delta(E_{\ph_0}(x))\bigr],
\]
which need not equal zero even when $\E_z[\Delta(z)]=0$, since the Jacobian-weighted expectation differs from the unweighted residual mean.

\begin{proof}
\textbf{Population identity.} Apply the tower property with respect to $(x_t,t)$:
\[
  \E_{z,t,\epsilon}[f^*(x_t,t)] = \E_{z,t,\epsilon}[\E[\epsilon-z\mid x_t,t]] = \E_{z,t,\epsilon}[\epsilon-z].
\]
This equals the right-hand side of $\E[\epsilon-z]$ inside $\Delta$, so $\E_z[\Delta(z)]=0$.

\textbf{Pointwise nonvanishing.} At fixed $z$ and $t>0$, $f^*(x_t,t)=\E[\epsilon'-z_0'\mid x_t,t]$ averages over all $(z_0',\epsilon')$ pairs whose interpolant lands at $x_t$, not just $(z,\epsilon)$. The posterior has nontrivial spread for $t>0$, so $\Delta_t(z)\neq 0$ in general.

\textbf{Slice-wise low-noise limit.} Fix $z\in\mathrm{supp}(p_z)$ with $p_z\in C^2$ and $p_z(z)>0$. The Bayes posterior is $p(z_0'\mid x_t,t)\propto p_z(z_0')\exp(-\|x_t-(1-t)z_0'\|^2/(2t^2))$, a Gaussian likelihood centered at $x_t/(1-t)$ with covariance $\Sigma_t=t^2/(1-t)^2 I$ tilted by the prior. The Laplace expansion at small $\|\Sigma_t\|$ gives
\begin{equation}\label{eq:laplace-mean}
  \E[z_0'\mid x_t,t] = \frac{x_t}{1-t} + \Sigma_t\,\nabla\log p_z\!\Bigl(\tfrac{x_t}{1-t}\Bigr) + O(\|\Sigma_t\|^2),
\end{equation}
uniformly on compact $K\subset\mathrm{supp}(p_z)$ by $p_z\in C^2(K)$ and $\inf_K p_z>0$. Substituting $x_t=(1-t)z+t\epsilon$, expanding $x_t/(1-t)=z+t\epsilon+t^2\epsilon+O(t^3)$ and $\Sigma_t=t^2 I+O(t^3)$, and Taylor of $\nabla\log p_z$ at $z$ gives $\E[z_0'\mid x_t,t]=z+t\epsilon+t^2\epsilon+t^2\nabla\log p_z(z)+O(t^3)$. Using $\epsilon'-z_0'=(x_t-z_0')/t$, $f^*(x_t,t)=-z-t\epsilon-t\nabla\log p_z(z)+O(t^2)$. Averaging over $\epsilon$ kills the $t\epsilon$ term, leaving
\[
  \Delta_t(z) = -t\,\nabla\log p_z(z) + O(t^2),
\]
which is $O(t)$ and vanishes pointwise on $\mathrm{supp}(p_z)$ as $t\to 0^+$. For $z\notin\mathrm{supp}(p_z)$, $\Delta_t(z)\not\to 0$ in general.
\end{proof}

\subsection{Information preservation and Pareto regime}\label{app:pareto}

\begin{assumption}\label{asm:smooth}
$\Lrec\in C^3$ on a neighbourhood $U$ of $\ph_0$, with $\nabla_\ph\Lrec(\ph_0)=0$ and Hessian $H\coloneqq\nabla_\ph^2\Lrec(\ph_0)\succeq\mu_{\mathrm{rec}}I\succ 0$.
\end{assumption}

\begin{remark}[Reading $H\succ 0$ in over-parameterized networks]\label{rem:H-quotient}
Strict positivity is a simplifying assumption. For over-parameterized encoders, $\Lrec$ has flat directions from reparameterization symmetries (permutation of hidden units, weight-norm rescaling, and similar things). The analysis should be read on the corresponding quotient, equivalently as $H\succ 0$ restricted to the orthogonal complement of the symmetry-group tangent at $\ph_0$.
\end{remark}

\begin{assumption}\label{asm:critic-smooth}
Both $\Lcsa(\,\cdot\,;\theta_0,w)$ and $\Lcrit(\,\cdot\,;\theta_0)$ are $C^2$ on $U$. Let $g\coloneqq\nabla_\ph\Lcsa(\ph_0;\theta_0,w)$ denote the stop-gradient pseudo-gradient. It is computed by autograd with $\vpred$ treated as constant in $\ph$. Assume $g\neq 0$.
\end{assumption}

\begin{remark}[$g$ is a vector field, not necessarily a scalar gradient]\label{rem:vector-field}
The pseudo-gradient $g$ is a vector field on $\Phi$. The implicit function theorem (IFT) framework does not require $g$ to be the gradient of a scalar function. It only needs $F(\ph,\lambda_c)=\nabla_\ph\Lrec(\ph)+\lambda_c\nabla_\ph\Lcsa(\ph;\theta_0,w)$ to be $C^1$ with invertible $\partial_\ph F$ at $(\ph_0,0)$. The trajectory $\ph^*(\lambda_c)$ in \Cref{thm:info-preservation} is well-defined regardless.
\end{remark}

\begin{theorem}[First-order Pareto improvement]\label{thm:info-preservation}
Under \Cref{asm:smooth,asm:critic-smooth}, let $\ph^*(\lambda_c)$ be the local stationary point of $\nabla_\ph\Lrec(\ph)+\lambda_c\nabla_\ph\Lcsa(\ph;\theta_0,w)=0$ near $\ph_0$. Throughout, $g$ denotes the stop-gradient pseudo-gradient of \Cref{asm:critic-smooth}. For all sufficiently small $\lambda_c>0$:
\begin{enumerate}[label=(\roman*),leftmargin=*,nosep]
\item $\ph^*(\lambda_c)=\ph_0-\lambda_c H^{-1}g+O(\lambda_c^2)$.
\item $\Lrec(\ph^*)-\Lrec(\ph_0)=\tfrac{\lambda_c^2}{2}g^\top H^{-1}g+O(\lambda_c^3)$.
\item $g^\top(\ph^*-\ph_0)=-\lambda_c g^\top H^{-1}g+O(\lambda_c^2)$.
\end{enumerate}
Since $g^\top H^{-1}g>0$, the surrogate progress-to-cost ratio $|g^\top(\ph^*-\ph_0)|/\Delta\Lrec=2/\lambda_c+O(1)\to\infty$ as $\lambda_c\to 0^+$.
\end{theorem}

\begin{proof}
Define $F:\Phi\times\R\to\Phi$ by
\[
  F(\ph,\lambda_c) \coloneqq \nabla_\ph\Lrec(\ph)+\lambda_c\,\nabla_\ph\Lcsa(\ph;\theta_0,w).
\]
$F$ is $C^1$ jointly in $(\ph,\lambda_c)$ from $\Lrec\in C^3$ (\Cref{asm:smooth}) and $\Lcsa\in C^2$ (\Cref{asm:critic-smooth}). Also $F(\ph_0,0)=0$ and $\partial_\ph F|_{(\ph_0,0)}=H\succ 0$ by \Cref{asm:smooth} (read on the symmetry quotient as in \Cref{rem:H-quotient}). The IFT then applies. It gives open $V\subset\R$ around $0$ and $U'\subset U$ around $\ph_0$. For every $\lambda_c\in V$, a unique $\ph^*(\lambda_c)\in U'$ satisfies $F(\ph^*(\lambda_c),\lambda_c)=0$. The map $\lambda_c\mapsto\ph^*(\lambda_c)$ is $C^1$. Implicit differentiation at $\lambda_c=0$ gives $\d\ph^*/\d\lambda_c|_0 = -H^{-1}g$, hence $\norm{\ph^*-\ph_0}=O(\lambda_c)$. To upgrade the remainder to $O(\lambda_c^2)$, expand the gradients in the stationarity equation $\nabla_\ph\Lrec(\ph^*)+\lambda_c\nabla_\ph\Lcsa(\ph^*;\theta_0,w)=0$ via Taylor at $\ph_0$. Using $\nabla_\ph\Lrec(\ph_0)=0$, $\nabla_\ph^2\Lrec(\ph_0)=H$, $\Lrec\in C^3$, and $\Lcsa\in C^2$,
\[
  H(\ph^*-\ph_0) + \lambda_c g
  = O(\norm{\ph^*-\ph_0}^2) + \lambda_c\,O(\norm{\ph^*-\ph_0})
  = O(\lambda_c^2),
\]
where the last equality uses $\norm{\ph^*-\ph_0}=O(\lambda_c)$. Inverting $H$,
\[
  \ph^*(\lambda_c) = \ph_0-\lambda_c H^{-1}g+O(\lambda_c^2),
\]
which is (i).

\textbf{Taylor for $\Lrec$.} Using $\nabla_\ph\Lrec(\ph_0)=0$, $\nabla_\ph^2\Lrec(\ph_0)=H$, and $\Lrec\in C^3$,
\[
  \Lrec(\ph^*) - \Lrec(\ph_0) = \tfrac{1}{2}(\ph^*-\ph_0)^\top H(\ph^*-\ph_0) + O(\norm{\ph^*-\ph_0}^3).
\]
Substituting $\ph^*-\ph_0=-\lambda_c H^{-1}g+O(\lambda_c^2)$ and using $HH^{-1}=I$ on the quadratic term gives $(\ph^*-\ph_0)^\top H(\ph^*-\ph_0)=\lambda_c^2 g^\top H^{-1}g+O(\lambda_c^3)$, and $\norm{\ph^*-\ph_0}^3=O(\lambda_c^3)$. Combining, (ii) holds with
\[
  \bigl|\Lrec(\ph^*)-\Lrec(\ph_0)-\tfrac{\lambda_c^2}{2}g^\top H^{-1}g\bigr|\leq C_{\mathrm{rec}}\,\lambda_c^3.
\]

\textbf{Surrogate pairing for (iii).} Taking the inner product of $\ph^*-\ph_0=-\lambda_c H^{-1}g+O(\lambda_c^2)$ with $g$ gives (iii):
\[
  \bigl|g^\top(\ph^*-\ph_0) + \lambda_c\,g^\top H^{-1}g\bigr|\leq C_{\mathrm{CSA}}\,\lambda_c^2.
\]

The constant $C_{\mathrm{rec}}>0$ depends on third derivatives of $\Lrec$ on a closed ball $\bar B(\ph_0,r)\subset U'$ that contains $\ph^*(\lambda_c)$ for all $\lambda_c\in[0,\lambda_c^{\max}]$, and $C_{\mathrm{CSA}}>0$ tracks the implicit constant in the trajectory expansion. $H\succ 0$ and $g\neq 0$ give $g^\top H^{-1}g>0$, and the surrogate progress-to-cost ratio follows from dividing (iii) by (ii).
\end{proof}

\begin{remark}[Reading of (iii)]\label{rem:pareto-iii}
Statement (iii) is the first-order pairing of the CSA direction $g$ with the trajectory displacement $\ph^*-\ph_0$. This is the surrogate progress that training reduces by construction. The corresponding decrease in $\Lcrit$ is bridged in \Cref{thm:diffusion-bound} through the alignment assumption \Cref{asm:align}. \Cref{lem:value-equiv} then carries it over to $\Ldiff$.
\end{remark}

\subsection{Diffusion loss improvement}\label{app:diff-loss}

\begin{assumption}\label{asm:opt}
$\theta_0=\argmin_\theta\Ldiff(\theta;\ph_0)$. The baseline diffusion model is optimal for the baseline encoder.
\end{assumption}

\begin{assumption}[Critic-CSA gradient alignment]\label{asm:align}
Let $g\coloneqq\nabla_\ph\Lcsa(\ph_0;\theta_0,w)$ be the stop-gradient pseudo-gradient that training optimizes. It is the gradient computed by autograd with $\sg(\vpred)$ treated as a constant in $\ph$, as in \Cref{prop:grad}. Let $g_{\mathrm{crit}}\coloneqq\nabla_\ph\Ldiff(\theta_0;\ph_0)$ be the true gradient of the conditional diffusion loss with respect to the encoder. We assume
\[
  g_{\mathrm{crit}}^\top H^{-1}g \;>\; 0.
\]
\end{assumption}

The descent direction $-H^{-1}g$ in \Cref{thm:info-preservation} comes from the stop-gradient pseudo-gradient. This is what training actually follows. The true gradient $g_{\mathrm{crit}}$ of the conditional diffusion loss is a different object. It includes a term involving the Jacobian of $f_{\theta_0}$ that the stop-gradient drops from the training signal. \Cref{asm:align} says that this dropped term does not flip the sign at $\ph_0$. So the descent direction used by training also descends the true $\Ldiff(\theta_0;\,\cdot\,)$ at $\ph_0$. We treat \Cref{asm:align} as a working assumption. Empirically the reconstruction quality stays flat while the downstream generation quality improves on motion and ImageNet, consistent with the Pareto regime predicted by the analysis.

\begin{remark}[Decomposition of the alignment condition]\label{rem:align-decomp}
Write $\delta\coloneqq g_{\mathrm{crit}}-g$. By construction $\delta$ collects the Jacobian term that $\sg$ removes from the training signal. It also captures the structural difference between $\Ldiff$ and $\Lcsa$. Then
\[
  g_{\mathrm{crit}}^\top H^{-1}g \;=\; \|g\|_{H^{-1}}^2 \;+\; \delta^\top H^{-1}g.
\]
The first term is non-negative and vanishes only at $g=0$. So \Cref{asm:align} is equivalent to $\delta^\top H^{-1}g > -\|g\|_{H^{-1}}^2$. A simple sufficient (but not necessary) condition is $\delta^\top H^{-1}g\geq 0$. A more testable condition is $\|\delta\|_{H^{-1}}\leq c\,\|g\|_{H^{-1}}$ with $c<1$. Cauchy-Schwarz then gives $|\delta^\top H^{-1}g|\leq c\|g\|_{H^{-1}}^2$. In standard diffusion architectures $f_{\theta_0}$ is locally Lipschitz with $\|J_{f_{\theta_0}}\|_{op}\leq L$ on $\mathrm{supp}(p_{x_t,t})$. This gives the per-instance bound $\|\delta\|\leq 2(1-t)\,L\,\|J_{E_\ph}\|_{op}\,\|r\|$. The same baseline-residual scale appears in $\|g\|$, so $\|\delta\|/\|g\|$ stays bounded. \Cref{asm:align} reduces to a non-extreme-anti-alignment condition. The ratio $\|\delta\|/\|g\|$ can be probed directly during fine-tuning. To probe the Euclidean alignment between $g$ and $g_{\mathrm{crit}}$, we compared their minibatch estimates on $128$ single-image batches from ImageNet~\cite{deng2009imagenet,imagenet15russakovsky} across two seeds, using the baseline encoder and a frozen critic. Their Euclidean inner product was positive on every batch (mean cosine $0.93$), and $\|\delta\|_2/\|g\|_2$ was below $1$ throughout (mean $0.40$).
\end{remark}

\begin{theorem}[Conditional diffusion-loss improvement under gradient alignment; formal form of \Cref{thm:diff-loss}]\label{thm:diffusion-bound}
Under \Cref{asm:smooth,asm:critic-smooth,asm:opt,asm:align} and $\lambda_c>0$ sufficiently small,
\[
  \min_\theta\Ldiff(\theta;\ph^*(\lambda_c)) < \min_\theta\Ldiff(\theta;\ph_0).
\]
\end{theorem}

\begin{proof}
Four-step chain.
\begin{enumerate}[label=(\roman*),leftmargin=*,nosep]
\item $\theta_0$ is fixed (the baseline of \Cref{asm:opt}). By definition of the minimum, $\min_{\theta'}\Ldiff(\theta';\ph^*)\leq\Ldiff(\theta_0;\ph^*)$.
\item By \Cref{lem:value-equiv}, $\Ldiff(\theta_0;\ph^*)=\Lcrit(\ph^*;\theta_0)$.
\item The IFT direction is the pseudo-gradient of $\Lcsa$. From \Cref{thm:info-preservation}, $\ph^*(\lambda_c)-\ph_0=-\lambda_c H^{-1}g+O(\lambda_c^2)$ with $g=\nabla_\ph\Lcsa(\ph_0;\theta_0,w)$ from \Cref{asm:critic-smooth}. This is what training follows, and we Taylor-expand $\Lcrit$, not $\Lcsa$. By \Cref{lem:value-equiv}, $\Lcrit(\,\cdot\,;\theta_0)$ and $\Ldiff(\theta_0;\,\cdot\,)$ agree as value functions of $\ph$. Their true gradients at $\ph_0$ both equal $g_{\mathrm{crit}}=\nabla_\ph\Ldiff(\theta_0;\ph_0)$. Substituting,
\[
  \Lcrit(\ph^*;\theta_0) - \Lcrit(\ph_0;\theta_0) = -\lambda_c\,g_{\mathrm{crit}}^\top H^{-1}g + O(\lambda_c^2).
\]
The form is mixed. $g$ comes from $\Lcsa$ as the training trajectory (\Cref{thm:info-preservation}). $g_{\mathrm{crit}}$ comes from $\Lcrit\equiv\Ldiff$ in value as the Taylor coefficient (\Cref{lem:value-equiv}). $\Lcsa$ never enters the Taylor expansion. \Cref{asm:align} assumes $g_{\mathrm{crit}}^\top H^{-1}g>0$. The difference is then strictly negative for all sufficiently small $\lambda_c>0$.
\item By \Cref{lem:value-equiv} and \Cref{asm:opt}, $\Lcrit(\ph_0;\theta_0)=\min_\theta\Ldiff(\theta;\ph_0)$.
\end{enumerate}
Concatenating gives the strict inequality.
\end{proof}

\begin{corollary}[Fixed-$\theta_0$ improvement]\label{cor:finetune}
Under \Cref{asm:smooth,asm:critic-smooth,asm:align} and $\lambda_c>0$ sufficiently small,
\[
  \Ldiff(\theta_0;\ph^*) \;<\; \Ldiff(\theta_0;\ph_0).
\]
Steps (i)--(iii) of the proof of \Cref{thm:diffusion-bound} do not invoke \Cref{asm:opt} and give this directly.
\end{corollary}

\begin{remark}[Fine-tuning inherits the bound]\label{rem:finetune}
At fixed $\theta_0$, the diffusion loss strictly decreases on the post-trained latents. No retraining of $\theta$ is required. \Cref{thm:diffusion-bound} adds \Cref{asm:opt} to upgrade this to a min-over-$\theta$ statement. Fine-tuning $\theta$ from $\theta_0$ on $\ph^*$ can only decrease the loss further. The practical fine-tune setting therefore inherits the strict bound either way.
\end{remark}

\begin{remark}[Approximate baseline optimality]\label{rem:approx-opt}
If $\Ldiff(\theta_0;\ph_0)=\min_\theta\Ldiff(\theta;\ph_0)+\epsilon_{\mathrm{opt}}$, then \Cref{thm:diffusion-bound} relaxes to
\[
  \min_\theta\Ldiff(\theta;\ph^*) \leq \min_\theta\Ldiff(\theta;\ph_0)+\epsilon_{\mathrm{opt}}-\lambda_c\,g_{\mathrm{crit}}^\top H^{-1}g+O(\lambda_c^2).
\]
The improvement persists as long as $\lambda_c\,g_{\mathrm{crit}}^\top H^{-1}g>\epsilon_{\mathrm{opt}}$.
\end{remark}

\subsection{Frequency-dependent effective temperature (DenseAM extension)}\label{app:freq-temp}

\citet{pham2025memorization} identify the empirical diffusion marginal
\[
  p(x_t,t)=\tfrac{1}{K}\sum_\mu (2\pi\sigma^2 t)^{-N/2}\exp\!\bigl(-\norm{x_t-\xi_\mu}^2/(2\sigma^2 t)\bigr)
\]
with a Dense Associative Memory at scalar inverse temperature $\beta(t)=1/(2\sigma^2 t)$. We extend this identification to a per-mode effective temperature under spatial stationarity.

\begin{proposition}[Per-mode effective temperature]\label{prop:freq-temp}
Assume the memories $\{\xi_\mu\}_{\mu=1}^K$ form a temporally wide-sense stationary (WSS) process: constant mean $\E[\xi_\mu(s)]=\mu_0$ for all $s$, and autocovariance $\mathrm{Cov}(\xi_\mu(s),\xi_\mu(t))$ depending only on the lag $|s-t|$. By Wiener-Khinchin, the population power spectral density (PSD) is the temporal Fourier transform of the autocovariance:
\[
  \PSD(k) \;\coloneqq\; \mathcal{F}\bigl[\mathrm{Cov}(\xi_\mu(0),\xi_\mu(\cdot))\bigr](k).
\]
The empirical estimator $\hat \PSD(k)\coloneqq K^{-1}\sum_{\mu=1}^K|(\FFT\xi_\mu)(k)|^2$ converges to $\PSD(k)$ as $K\to\infty$ under ergodicity, with $O(K^{-1/2})$ deviation at finite $K$.

Because $\FFT$ is unitary, Parseval gives $\norm{x_t-\xi_\mu}^2=\sum_k|\hat x_{t,k}-\hat\xi_{\mu,k}|^2$, and each Boltzmann factor splits across modes:
\[
  p(x_t,t)
  = \frac{1}{K}\sum_\mu(2\pi\sigma^2 t)^{-N/2}\prod_k\exp\!\Bigl(-\frac{|\hat x_{t,k}-\hat\xi_{\mu,k}|^2}{2\sigma^2 t}\Bigr).
\]
The sum over $\mu$ does not factor across $k$. WSS gives second-order spectral uncorrelatedness across modes, not joint independence. We treat modes as independent as a working approximation. The per-mode signal and noise energies are then $E_{\mathrm{sig}}(k)=\PSD(k)$ and $E_{\mathrm{noise}}(k,t)=\sigma^2 t$, giving a per-mode SNR:
\begin{equation}\label{eq:beta-mode}
  \beta_{\mathrm{eff}}(k,t)
  \;\propto\; E_{\mathrm{sig}}(k)/E_{\mathrm{noise}}(k,t)
  \;=\; \PSD(k)/(\sigma^2 t).
\end{equation}
\end{proposition}

\begin{proof}
Apply the scalar identification of \citet{pham2025memorization} per mode to each Boltzmann factor above. The per-mode signal energy scales as $\PSD(k)$ by the Wiener-Khinchin definition.
\end{proof}

\begin{remark}[Reading of \Cref{prop:freq-temp}]\label{rem:freq-temp-scope}
The mode-independence step is a working approximation, not an exact identity. WSS gives only second-order uncorrelatedness across modes. Joint independence is a separate assumption that we adopt for tractability. The polynomial weight $w(k)=1+\kappa d(k)^p$ should be read as motivated by this approximation, not derived as a unique identity. The empirical spectral profile in Fig.~\ref{fig1a} supports this choice in our deployment regime.
\end{remark}

\begin{remark}[Static weight as SNR compensation]\label{rem:weight-snr}
For a power-law PSD $\PSD(k)\propto\norm{k}^{-\gamma}$, the SNR-compensation choice from \eqref{eq:beta-mode} is $\tilde w(k)\propto\beta_{\mathrm{eff}}^{-1}\propto\norm{k}^\gamma$. This makes $\tilde w(k)\beta_{\mathrm{eff}}(k,t)$ approximately constant across $k$. The encoder pressure $\tfrac{2}{N}\tilde w(k)|\hat r(k)|$ from \Cref{thm:spectral-grad} then applies more force at modes with smaller per-mode SNR. For $D>1$ the same construction applies axis-wise. The additive 1 in $w(k)=1+\kappa\,d(k)^p$ does not come from \eqref{eq:beta-mode}. It keeps $\tilde w(k)>0$, which \Cref{thm:strong-convex} uses. It also keeps a baseline pressure on low-frequency modes.
\end{remark}

\subsection{Practical considerations for text-to-motion deployment}\label{app:cond-extension}

The setup in \Cref{app:setup} is already text-conditioned. This subsection collects practical remarks for deployment.

\begin{remark}[Scope of the improvement]\label{rem:cond-scope}
\Cref{thm:diffusion-bound} states $\min_\theta\Ldiff(\theta;\ph^*) < \min_\theta\Ldiff(\theta;\ph_0)$ where $\Ldiff$ is the joint-expectation loss $\E_{(x,y),t,\epsilon}[\cdot]$ (equivalently $\E_{y\sim p_y}[\Ldiff^y]$ via tower property, with $\Ldiff^y$ the per-prompt loss). The result is a min-over-$\theta$ statement about the prompt-averaged loss. It does not say a fixed $\theta$ achieves this drop, nor does it say each fixed prompt $y$ improves. Text alignment, R-Precision, FID, and related metrics depend on the full sampler and decoder pipeline, so the theorem does not predict them.
\end{remark}

\begin{remark}[Classifier-free guidance and approximate optimality]\label{rem:cond-cfg}
Pretrained text-to-motion diffusion models are usually trained with classifier-free guidance (CFG)~\citep{ho2022classifier}. With probability $p_{\mathrm{uncond}}$ the text $y$ is replaced by a null token during training, so the actual training objective is a mixture of conditional and unconditional losses. Hence $\theta_0$ does not exactly minimize the conditional $\Ldiff(\theta;\ph_0)$ in \Cref{asm:opt}. \Cref{rem:approx-opt} covers this case. The improvement of \Cref{thm:diffusion-bound} persists when the conditional optimality gap is small relative to $\lambda_c\,g_{\mathrm{crit}}^\top H^{-1}g$. CSA uses the raw conditional prediction $f_{\theta_0}(x_t,t,y)$, not the CFG-guided combination used at sampling time.
\end{remark}

\begin{remark}[Encoder stays text-free]\label{rem:cond-encoder}
The conditioning on $y$ comes from the data distribution $p(x|y)$, not from the encoder. The map $E_\ph(x)$ takes only motion. Different texts induce different latent distributions because they select different motions, and the critic gradient picks up the text signal through $f_{\theta_0}(\cdot,\cdot,y)$.
\end{remark}

\section{ImageNet generalization study}\label{app:imagenet-csa}

We apply the CSA pipeline to class-conditional ImageNet~\cite{deng2009imagenet,imagenet15russakovsky}. The autoencoder is the Stable Diffusion VAE (SD-VAE) backbone~\cite{rombach2022high}, fine-tuned with CSA at $\kappa=1.0$, $\lambda_c=0.1$, $p=2.0$, for $5{,}000$ fine-tune steps with $1{,}000$-step warmup on $\lambda_c$, learning rate $10^{-6}$, batch size $8$. The diffusion model is a $200$M-parameter class-conditional UNet at $256{\times}256$ resolution. The architecture has channels $(192, 384, 576, 768)$, three layers per block, dropout $0.1$, and $1{,}001$ class embeddings. We train from scratch on the $1.28$M ImageNet training images for $300{,}000$ steps at batch size $128$. Peak learning rate is $10^{-4}$ with linear warmup over $5{,}000$ steps followed by cosine decay. We use exponential moving average (EMA) decay $0.9999$ and CFG dropout probability $0.10$. Reconstruction FID is measured on $50{,}000$ validation images. Generation FID uses $50$ samples per class generated via $50$-step deterministic sampling at CFG scale $w=2.0$. We use a single seed ($42$).

\begin{figure}[h]
\centering
\includegraphics[width=0.8\linewidth]{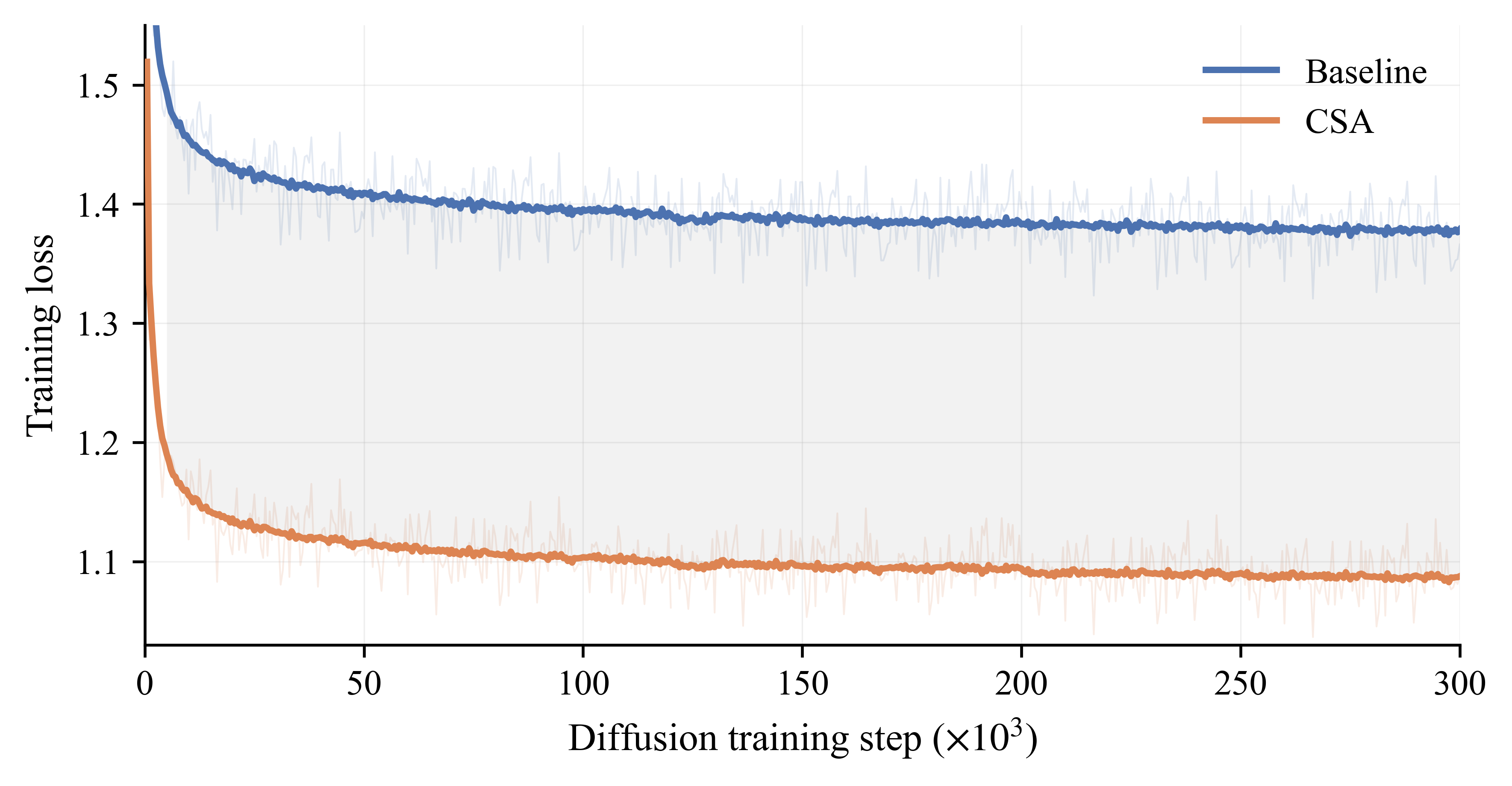}
\caption{Diffusion training loss on class-conditional ImageNet latents. CSA-fine-tuned latents (orange) yield consistently lower training loss than the SD-VAE baseline (blue) from step $1{,}000$ onward. Faded curves show raw step-wise loss. Solid curves show $0.999$-EMA-smoothed loss. The shaded band marks the sustained gap of about $0.30$.}
\label{fig:imagenet-loss}
\end{figure}

\Cref{tab:imagenet-csa} reports the paired baseline-vs-CSA result. Three observations stand out. Reconstruction FID stays close to baseline ($\Delta = +0.08$), and generation FID improves at all three checkpoints we evaluate ($\Delta = -2.24$ at $100{,}000$ training steps, $\Delta = -1.69$ at $200{,}000$ training steps, and $\Delta = -1.35$ at $300{,}000$ training steps). Moreover, the diffusion training loss curves separate by about $0.30$ from step $1{,}000$ onward (\Cref{fig:imagenet-loss}). Taken together, all three are consistent with the prediction of \Cref{thm:diff-loss}. The post-trained latent is easier for the diffusion to fit, and reconstruction is preserved at first order in $\lambda_c$.

\begin{table}[h]
\caption{ImageNet generalization study. Single seed ($42$), $200$M class-conditional UNet, CFG scale $w=2.0$. Reconstruction FID is computed on $50{,}000$ validation images. Generation FID is computed from $50{,}000$ CFG samples ($50$ per class).}
\label{tab:imagenet-csa}
\centering
\begin{tabular}{lccc}
\toprule
Metric & Baseline & CSA & $\Delta$ \\
\midrule
Reconstruction FID & $2.79$ & $\mathbf{2.87}$ & $+0.08$ \\
Generation FID at $100$k steps & $36.29$ & $\mathbf{34.05}$ & $-2.24$ \\
Generation FID at $200$k steps & $25.46$ & $\mathbf{23.77}$ & $-1.69$ \\
Generation FID at $300$k steps & $20.82$ & $\mathbf{19.47}$ & $-1.35$ \\
\bottomrule
\end{tabular}
\end{table}

\section{MeanFlow Identity Derivation}\label{app:meanflow-identity}

This appendix derives the JVP form of the average velocity~\eqref{eq:meanflow-jvp} used in Section~\ref{sec:Alpha-Flow} from the integral definition~\eqref{eq:meanflow-avg}, following~\citet{geng2025mean}.

Recall the average velocity definition:
\begin{equation}\label{eq:meanflow-avg-recall}
u(x_t, r_d, t, z) = \frac{1}{t-r_d}\int_{r_d}^t v(x_\tau, \tau, z)\,d\tau.
\end{equation}
Multiplying both sides by $(t-r_d)$ removes the prefactor and gives an integral identity:
\begin{equation}\label{eq:meanflow-multiplied}
(t-r_d)\,u(x_t, r_d, t, z) = \int_{r_d}^t v(x_\tau, \tau, z)\,d\tau.
\end{equation}
We now differentiate both sides of~\eqref{eq:meanflow-multiplied} with respect to $t$, treating $r_d$ as independent of $t$.

The right-hand side reduces by the fundamental theorem of calculus to
\begin{equation}\label{eq:meanflow-rhs}
\frac{d}{dt}\!\int_{r_d}^t v(x_\tau, \tau, z)\,d\tau = v(x_t, t, z).
\end{equation}
The left-hand side expands by the product rule:
\begin{equation}\label{eq:meanflow-lhs}
\frac{d}{dt}\bigl[(t-r_d)\,u(x_t, r_d, t, z)\bigr]
= u(x_t, r_d, t, z) + (t-r_d)\,\frac{d}{dt}u(x_t, r_d, t, z).
\end{equation}
Equating~\eqref{eq:meanflow-rhs} and~\eqref{eq:meanflow-lhs} and rearranging recovers the MeanFlow identity:
\begin{equation}\label{eq:meanflow-identity-derived}
u(x_t, r_d, t, z) = v(x_t, t, z) - (t-r_d)\,\frac{d}{dt}u(x_t, r_d, t, z).
\end{equation}
This matches~\eqref{eq:meanflow-jvp} in the main text.

At training time, the total derivative $\tfrac{d}{dt}u(x_t, r_d, t, z)$ is computed via the Jacobian-vector product (JVP) of the decoder network with the tangent vector $(v(x_t,t,z),\,0,\,1,\,0)$ along the $(x, r_d, t, z)$ axes. The instance of $u$ inside the time derivative is detached, denoted $u_{\psi-}$ in~\eqref{eq:meanflow-jvp}, to prevent recursive backpropagation through the network's own prediction~\citep{geng2025mean}. Without this stop-gradient, the loss would induce double backpropagation through the JVP and require higher-order optimization.

\section{Generation Metric}
\label{metric_app}

\noindent\textbf{Frechet Inception Distance (FID)}
The FID~\cite{heusel2017gans} measures the distribution distance between the generated and real interaction features.
\[
\text{FID} = \|\mu_{\text{gt}} - \mu_{\text{pred}}\|^2 + \text{Tr}(\Sigma_{\text{gt}} + \Sigma_{\text{pred}} - 2(\Sigma_{\text{gt}}\Sigma_{\text{pred}})^{\frac{1}{2}})
\]
where $\mu_{\text{gt}}$ and $\mu_{\text{pred}}$ are the mean ground-truth and generated interaction features, and $\Sigma$ represents the covariance matrix.

\noindent\textbf{Multimodal Distance (MM-Dist)}
This metric calculates the average Euclidean distance between each text feature and the generated interaction feature.
\[
\text{MM-Dist} = \frac{1}{N} \sum_{i=1}^{N} \|f_{t,i} - f_{m,i}\|
\]
where $f_{t,i}$ and $f_{m,i}$ are the features of the $i$th text-interaction pair.

\noindent\textbf{Diversity}
All generated interactions are randomly sampled to calculate the average Euclidean distance between two subsets.
\[
\text{Diversity} = \frac{1}{X_d} \sum_{i=1}^{X_d} \|x_i - x'_i\|
\]

\noindent\textbf{Multimodality (MModality)}
This metric assesses the variability given multiple text descriptions by calculating the average pairwise Euclidean distance between motion features.
\[
\text{MModality} = \frac{1}{J_m \times X_m} \sum_{j=1}^{J_m} \sum_{i=1}^{X_m} \|x_{j,i} - x'_{j,i}\|
\]
where $x_{j,i}$ and $x'_{j,i}$ are the features of the $j$th pair of the $i$th text description.

\newpage

\section{Implementation Details}
\label{imple_app}

\paragraph{Diffusion backbone Training.}
In DiMoT, we train the diffusion backbone for different tokenizers in a fixed way.
Firstly, each motion sequence is encoded into a set of $16$ latent tokens with a channel dimension of $32$, and the diffusion backbone denoises them conditioned on text. 
The denoiser is implemented as a transformer with an encoder-only topology. 
Each latent token is first projected to a $512$-dimensional hidden space, processed by an $11$-layer skip-connected Transformer encoder with $8$ attention heads, a feed-forward dimension of $1024$, GELU activations, and dropout of $0.1$, and then projected back to the original latent dimension. 
We use learned 1D positional embeddings for the latent token sequence. Conditioning is injected by prepending two condition tokens to the latent sequence: (1) a timestep token obtained from a sinusoidal timestep embedding followed by an MLP projection, and (2) a text token produced by a frozen CLIP text encoder followed by a 2-layer Transformer Encoder for feature refinement. 
Thus, the denoiser processes a joint sequence consisting of $16$ motion latent tokens, one timestep token, and one global text token. 
Notably, the Transformer backbone is not a plain stacked encoder: it adopts a skip-encoder design, in which hidden states from the early encoder blocks are concatenated with their symmetric late blocks via learnable linear fusion, thereby improving cross-layer information propagation while preserving global self-attention.
By default, the model is trained with a flow-matching velocity prediction objective and sampled with classifier-free guidance at guidance scale $3.0$ in 10 steps.
The training process uses a batch size of 192, a peak learning rate of 1e-4, a 1K-iteration linear warm-up, and a cosine decay schedule.
In the CSA stage, we use the same denoiser backbone, which remains frozen.


\paragraph{Tokenizer Training.}
We reproduce MLD++~\cite{motionlcm-v2} and adopt it as our baseline with multi-latent-token training. The baseline tokenizer is trained for 6K epochs with a peak learning rate of $1\times 10^{-4}$, a batch size of 96, and a cosine decay schedule following a 1K-epoch linear warmup. 
During the CSA stage, we train the encoder for 1K epochs and gradually introduce spectral reweighting over a 200-epoch warmup. The encoder is optimized with an $\ell_2$ reconstruction objective, and the CSA loss is weighted by $\lambda_c=0.1$. We evaluate CSA under both training-from-scratch and baseline fine-tuning settings. For the fine-tuning variant, we additionally study an encoder-side adapter implemented as a 2-layer residual block. The spectral hyperparameters are set to $\kappa=1.0$ and $p=2.0$.
%
%
The critic is pretrained on the baseline tokenizer's latents and remains frozen throughout CSA.
After CSA, we freeze the adapted encoder and train a separate generative diffusion model from scratch on its latents, using the same architecture and training configuration as the baseline diffusion model.
This generative model supplies diffusion-generated latent conditions for AMD training.
In detail, we replace deterministic decoding with a latent-conditioned generative decoder. The decoder uses a 9-layer Transformer~\cite{vaswani2017attention}, while the latent-to-frame upsampling projection is implemented as a 2-layer Transformer. We train the AMD decoder from scratch for 2K epochs, where latent conditions are sampled from the pretrained diffusion output with probability 0.25 and from the encoder output with probability 0.75. For 50\% of each batch, we apply marginal flow-matching supervision by setting $t=r_d$. 
For the remaining samples, we use an $\alpha$-curriculum that linearly decreases the generative mixing factor from 1.0 to 0.005 over the first 1K epochs and then sets it to 0 for the final 1K epochs. We further apply single-step classifier-free guidance~\cite{geng2025mean} with guidance scale 3.

\paragraph{Compute resources.}
All experiments were conducted on the High-Performance Computing cluster. We use a Linux server equipped with a 30-core Intel Xeon 6767P CPU, 235\,GiB of system memory, and an NVIDIA H200 GPU. 

\newpage
\section{AMD ablation}
\label{AMD_eval}

\begin{table*}[!h]
  \caption{AMD ablation.}
  \label{tab:amd_core_ablation}
  \centering
  \resizebox{0.9\textwidth}{!}{
  \begin{tabular}{lcccc|ccccc}
    \toprule
    \multirow{2}{*}{Variant} & \multirow{2}{*}{Proj.} & \multirow{2}{*}{Step} & \multirow{2}{*}{Target} & \multirow{2}{*}{$\alpha$ Sched. (Ratio)} & \multicolumn{5}{c}{Evaluation Metrics} \\
    \cmidrule(lr){6-10}
    & & & & & FID$\downarrow$ & R-Top3$\uparrow$ & MPJPE$\downarrow$ & Foot Sliding$\downarrow$ & AITS$\downarrow$ \\
    \midrule
    MLD++ (Baseline)      & --            & --     & --    & --              & 5.046 & 0.688 & 49.26 & 0.0186 & 1.05 \\
    \midrule
    \rowcolor[gray]{.95} \multicolumn{10}{l}{\textit{Target $\boldsymbol{v}$ Flow Matching Velocity Baseline}} \\
    V1: + Interpo.     & Interpolation & 1 NFE  & $v$   & None (0\%)      & 7.625 & 0.551 & 66.12 & 0.0255 & 3.05\\
    V2: + Trans.    & Transformer   & 16 NFE & $v$   & None (0\%)      & 4.951 & 0.708 & 47.12 & 0.0181 & 45.16 \\
    V3: + Trans.    & Transformer   & 1 NFE  & $v$   & None (0\%)      & 6.842 & 0.589 & 58.45 & 0.0221 & 4.82 \\
    \midrule
    \rowcolor[gray]{.95} \multicolumn{10}{l}{\textit{Target $\boldsymbol{u}$ Projection Search}} \\
    V4: Interpolation & Interpolation & 1 NFE  & $u$   & None (0\%)      & 5.215 & 0.675 & 49.85 & 0.0191 & 3.13 \\ 
    V5: MLP           & MLP           & 1 NFE  & $u$   & None (0\%)      & 5.012 & 0.694 & 47.53 & 0.0178 & 3.64 \\ 
    \midrule
    \rowcolor[gray]{.95} \multicolumn{10}{l}{\textit{Target $\boldsymbol{u}$ + Marginal Distribution ratio $t=r$}} \\
    V6: Ratio 25\%    & Transformer   & 1 NFE  & $u$   & Annealed (25\%) & 4.832 & 0.710 & 46.90 & 0.0183 & 4.80 \\ 
    \rowcolor{blue!15}V7: Ratio 50\% (Default)    & Transformer   & 1 NFE  & $u$   & Annealed (50\%) & 4.781 & 0.716 & 46.76 & 0.0175 & 4.80 \\
    V8: Ratio 75\%    & Transformer   & 1 NFE  & $u$   & Annealed (75\%) & 4.805 & 0.714 & 46.82 & 0.0176 & 4.80 \\ 
    \midrule
    \bottomrule
  \end{tabular}}
\end{table*}

\section{CSA ablation}
\begin{figure}[h]
    \centering
    \includegraphics[width=0.9\linewidth]{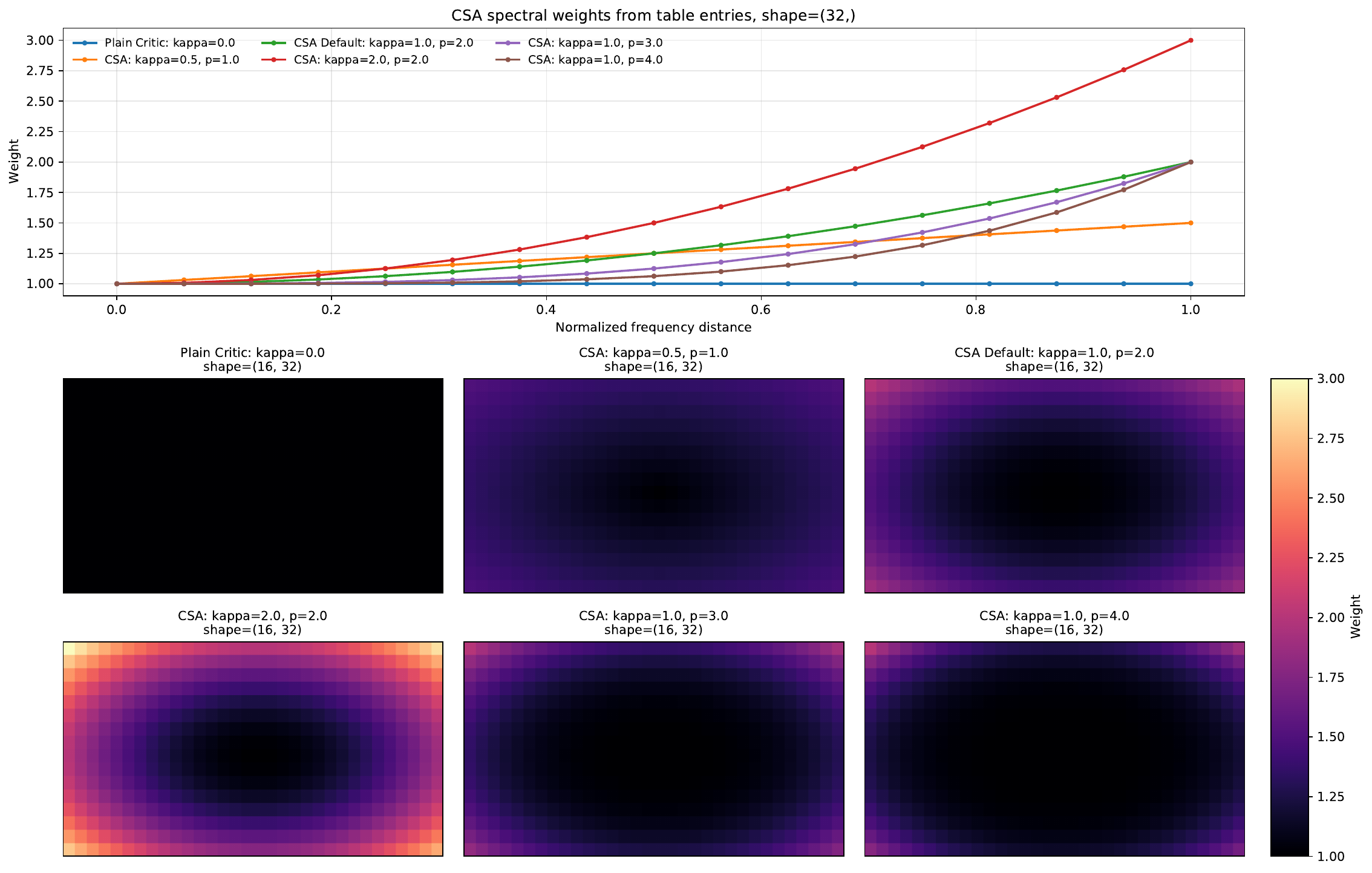}
    \caption{Spectral Reweighting Curve.}
    \label{fig:placeholder}
\end{figure}


\label{csa_results}

\begin{table}[h]
\centering
\caption{Ablation of CSA Spectral Weights ($\kappa, p$) and Critic Scale ($\lambda_c$)}
\label{tab:csa_ablation}
\resizebox{0.8\textwidth}{!}{
\begin{tabular}{lcccccccc}
\toprule
Method & $\kappa$ & $p$ & $\lambda_{c}$ & rFID $\downarrow$ & MPJPE $\downarrow$  & FID $\downarrow$ & R-Precision $\uparrow$ \\
\midrule
Baseline & - & - & 0.0 & 0.080 & 49.26 & 5.046 & 0.688 \\
Plain Critic & 0.0 & - & 0.1 & 0.082 & 48.50 & 5.012 & 0.708 \\
\midrule
CSA & 0.5 & 1.0 & 0.1 & 0.084 & 47.85 & 4.985 & 0.728 \\
\rowcolor{blue!15} \textbf{CSA (Default)} & \textbf{1.0} & \textbf{2.0} & \textbf{0.1} & \textbf{0.086} & \textbf{47.18} & \textbf{4.967} & \textbf{0.746} \\
CSA & 2.0 & 2.0 & 0.1 & 0.125 & 51.32 & 5.215 & 0.712 \\
CSA & 1.0 & 3.0 & 0.1 & 0.188 & 58.45 & 5.682 & 0.665 \\
CSA & 1.0 & 4.0 & 0.1 & 0.245 & 66.80 & 5.932 & 0.621 \\
\midrule
CSA & 1.0 & 2.0 & 0.01 & 0.081 & 48.60 & 5.035 & 0.698 \\
CSA & 1.0 & 2.0 & 0.5 & 0.142 & 54.20 & 5.410 & 0.685 \\
\bottomrule
\end{tabular}}
\end{table}

\subsection{Effect of the Alpha Curriculum Schedule}
\label{app:alpha-schedule}

We compare the default linear curriculum with an AlphaFlow-style sigmoid curriculum~\cite{zhang2025alphaflow}, keeping the remaining AMD settings unchanged.
AMD is trained for 2K epochs.
For samples with non-degenerate intervals ($r_d<t$), both schedules decrease $\alpha$ from $1$ to $\alpha_{\min}=0.005$ over the first 1K epochs, followed by the exact MeanFlow objective ($\alpha=0$) for the final 1K epochs.
The degenerate-interval ratio remains $50\%$ for both schedules.

Let $e$ denote training progress in epochs and $q=e/1000$.
For schedule $S$, we use
\begin{equation}
\alpha_S(e)=
\begin{cases}
\alpha_{\min}+(1-\alpha_{\min})g_S(q),
& 0\leq e\leq 1000,\\
0, & 1000<e\leq 2000.
\end{cases}
\end{equation}
The linear schedule uses $g_{\mathrm{lin}}(q)=1-q$.
For the sigmoid schedule, with $\sigma(a)=(1+\exp(-a))^{-1}$ and $\gamma=25$, we use
\begin{equation}
g_{\mathrm{sig}}(q)=
\frac{\sigma(\gamma/2)-\sigma(\gamma(q-1/2))}
{\sigma(\gamma/2)-\sigma(-\gamma/2)}.
\end{equation}
This normalization gives the same initial and final $\alpha$ values for both curricula.

\begin{table}[htbp]
    \centering
    \caption{Comparison of $\alpha$ curriculum schedules.
    Both schedules use the same training duration and endpoint values.}
    \label{tab:alpha-schedule}
    \begin{tabular}{lcccc}
        \toprule
        Schedule & FID$\downarrow$ & R-Top3$\uparrow$
        & MPJPE$\downarrow$ & Foot Sliding$\downarrow$ \\
        \midrule
        Linear (default) & 4.781 & 0.716 & 46.76 & 0.0175 \\
        Sigmoid ($\gamma=25$) & 4.856 & 0.687 & 49.52 & 0.0178 \\
        \bottomrule
    \end{tabular}
\end{table}

The linear schedule performs better on all four reported metrics in this comparison.
The sigmoid schedule concentrates the transition around the middle of the first training stage, whereas the linear schedule changes $\alpha$ at a constant rate.
These results support the default linear curriculum under the evaluated setting.

\section{User Study against InterMask}
\label{app:intermask-user-study}

We conduct a user study comparing DiMoT with InterMask~\cite{javed2024intermask}.
The study involves 11 participants and 20 randomly selected samples.
We evaluate preferences along four dimensions: text alignment, physical realism, interaction quality, and overall quality.

\begin{table}[htbp]
    \centering
    \caption{User preferences for InterMask and DiMoT.
    Higher percentages indicate greater preference.
    Ties are excluded from the overall-quality comparison.}
    \label{tab:intermask-user-study}
    \begin{tabular}{lcc}
        \toprule
        Criterion & InterMask & DiMoT \\
        \midrule
        Text alignment   & 53.33\% & 46.67\% \\
        Physical realism & 46.15\% & 53.85\% \\
        Interaction quality & 43.80\% & 56.20\% \\
        Overall quality  & 47.39\% & 52.61\% \\
        \bottomrule
    \end{tabular}
\end{table}

DiMoT receives higher preference percentages for physical realism, interaction quality, and overall quality, while InterMask receives higher preference for text alignment.
The overall-quality difference is modest, and the study is limited in participant and sample size.
We therefore interpret these results as descriptive evidence of user preference in this comparison, rather than as a statistically established advantage.

\newpage



\end{document}